\documentclass{article}

\usepackage{microtype}
\usepackage{graphicx}
\usepackage{subcaption}
\usepackage{xcolor}
\usepackage{multirow}
\usepackage{bm}
\usepackage{caption}
\usepackage{tabularx}
\usepackage{diagbox}
\usepackage{CJKutf8}
\usepackage{booktabs} % for professional tables
\usepackage{adjustbox}
\usepackage{rotating}
\usepackage{floatrow}

\usepackage{hyperref}

\usepackage[preprint]{icml2026}

\usepackage{amsmath}
\usepackage{amssymb}
\usepackage{mathtools}
\usepackage{amsthm}

\usepackage[capitalize,noabbrev]{cleveref}

\makeatletter
\newcommand{\precneq}{\mathrel{\text{\prec@eq}}}
\newcommand{\prec@eq}{%
  \oalign{%
    \hidewidth$\m@th\prec$\hidewidth\cr
    \noalign{\nointerlineskip\kern1ex}%
    $\m@th\smash{\Neq}$\cr
    \noalign{\nointerlineskip\kern-.5ex}%
  }%
}
\newcommand{\Neq}{\raisebox{0.65ex}{\rotatebox{90}{\scalebox{1}[-1]{$\nshortmid$}}}}
\makeatother

\theoremstyle{plain}

\theoremstyle{definition}

\theoremstyle{remark}

\usepackage[textsize=tiny]{todonotes}

\icmltitlerunning{BAPS: A Fine-Grained Low-Precision Scheme for Softmax in Attention via Block-Aware Precision reScaling}

\begin{document}

\twocolumn[
\icmltitle{BAPS: A Fine-Grained Low-Precision Scheme for Softmax in Attention \\ via Block-Aware Precision reScaling}

  % It is OKAY to include author information, even for blind submissions: the
  % style file will automatically remove it for you unless you've provided
  % the [accepted] option to the icml2026 package.

  % List of affiliations: The first argument should be a (short) identifier you
  % will use later to specify author affiliations Academic affiliations
  % should list Department, University, City, Region, Country Industry
  % affiliations should list Company, City, Region, Country

  % You can specify symbols, otherwise they are numbered in order. Ideally, you
  % should not use this facility. Affiliations will be numbered in order of
  % appearance and this is the preferred way.
\icmlsetsymbol{equal}{*}

\begin{icmlauthorlist}
\icmlauthor{Zisheng Ye}{equal,yyy}
\icmlauthor{Xiaoyu He}{equal,yyy}
\icmlauthor{Maoyuan Song}{yyy}
\icmlauthor{Guoliang Qiu}{yyy}
\icmlauthor{Chao Liao}{yyy}
\icmlauthor{Chen Wu}{yyy}
\icmlauthor{Yonggang Sun}{comp}
\icmlauthor{Zhichun Li}{comp}
\icmlauthor{Xiaoru Xie}{comp}
\icmlauthor{Yuanyong Luo}{comp}
\icmlauthor{Hu Liu}{comp}
\icmlauthor{Pinyan Lu}{yyy,sch}
\icmlauthor{Heng Liao}{comp}
%\icmlauthor{}{sch}
%\icmlauthor{Firstname8 Lastname8}{sch}
%\icmlauthor{Firstname8 Lastname8}{yyy,comp}
%\icmlauthor{}{sch}
%\icmlauthor{}{sch}
\end{icmlauthorlist}

  \icmlaffiliation{yyy}{Taylor Lab, Huawei}
  \icmlaffiliation{comp}{HiSilicon, Huawei}
  \icmlaffiliation{sch}{Key Laboratory of Interdisciplinary Research of Computation and Economics, Shanghai University of Finance and
Economics, Shanghai, China}

\icmlcorrespondingauthor{Pinyan Lu}{lu.pinyan@mail.shufe.edu.cn}
\icmlcorrespondingauthor{Heng Liao}{liao.heng@hisilicon.com}

% You may provide any keywords that you
% find helpful for describing your paper; these are used to populate
% the "keywords" metadata in the PDF but will not be shown in the document
\icmlkeywords{low precision arithmetic, softmax function, hardware software co-design, inference acceleration}

\vskip 0.3in
]

% this must go after the closing bracket ] following \twocolumn[ ...

% This command actually creates the footnote in the first column
% listing the affiliations and the copyright notice.
% The command takes one argument, which is text to display at the start of the footnote.
% The \icmlEqualContribution command is standard text for equal contribution.
% Remove it (just {}) if you do not need this facility.

%\printAffiliationsAndNotice{}  % leave blank if no need to mention equal contribution
\printAffiliationsAndNotice{\icmlEqualContribution}
% otherwise use the standard text.

\begin{abstract}
As the performance gains from accelerating quantized matrix multiplication plateau, the softmax operation becomes the critical bottleneck in Transformer inference. This bottleneck stems from two hardware limitations: (1) limited data bandwidth between matrix and vector compute cores, and (2) the significant area cost of high-precision (FP32/16) exponentiation units (EXP2). To address these issues, we introduce a novel low-precision workflow that employs a specific 8-bit floating-point format (HiF8) and block-aware precision rescaling for softmax.
Crucially, our algorithmic innovations make low-precision softmax feasible without the significant model accuracy loss that hampers direct low-precision approaches.
Specifically, our design (i) halves the required data movement bandwidth by enabling matrix multiplication outputs constrained to 8-bit, and (ii) substantially reduces the EXP2 unit area by computing exponentiations in low (8-bit) precision. Extensive evaluation on language models and multi-modal models confirms the validity of our method. By alleviating the vector computation bottleneck, our work paves the way for doubling end-to-end inference throughput without increasing chip area, and offers a concrete co-design path for future low-precision hardware and software.

\end{abstract}

\section{Introduction}
\label{sec:intro}

% to be continued. the current flow of the intro seems strange.

% \qgl{first hightlight the importance of acceleration.}

% \qgl{second, list the challenge of the acceleration and introduce the existing works according to their direction}

% \qgl{third, emphasize the gap of the existing works}

% \qgl{finally, state the novelty of our contribution and the structure of the paper}

Modern accelerators typically comprise two types of components. The first consists of general-purpose accelerators such as streaming multi-processors on GPU and Vector Cores on NPU \cite{Liao2019DaVinci}, which contain a lot of arithmetic logic units (ALUs). The second type includes special-purpose accelerators, designed with domain-specific architectures (DSAs) to efficiently execute a small range of highly-demanded tasks. Since general matrix multiplications (GEMMs) are fundamental to a variety of diverse fields—including machine learning \cite{bishop2006pattern}, computer vision \cite{foundationsCVbook}, robotics \cite{murray1994mathematical}, and scientific computing \cite{Dongarra1998}, such operations are commonly assigned onto special-purpose components for best performance. On NPUs, these are known as Cube Cores, while GPUs often employ Tensor Cores for such operations.  

The advent of Transformer \cite{Ashish2017Attention} has driven remarkable progress across artificial intelligence generated content (AIGC), powering breakthroughs in natural language processing and reasoning \citep{openai2024gpt4, hu2024case}, video generation \cite{peebles2023scalable}, agent workflows \cite{Wang2024Survey}, and robotic controls \cite{sanghai2024advances}. The most computationally intensive component in Transformer is the multi-head attention mechanism, which takes GEMMs and softmax normalizations as its technical cornerstone. To alleviate GEMMs bottlenecks, quantization \cite{Itay2018Quantized} has become a fundamental technique, reducing memory bandwidth and enabling efficient computation on dedicated hardware such as Nvidia Tensor Cores and Huawei Cube Cores. An alternative approach, BLASST \cite{yuan2025blasst}, has recently been proposed as a skip scheme to avoid the expensive computational cost paid on softmax computation. Attempts at integrating these two approaches have however been met with the hurdle of precision conversion: BLASST is a fundamentally independent technique from quantized GEMMs, that starts by converting the low-precision outputs from the previous stage back to high-precision at an additional overhead, and does not take full advantage of the reduced computation cost that quantization and reduced precision allows. Despite these observations, very few efforts have ever been placed on directly reducing the cost of softmax computation via low-precision arithmetic.

%In this paper, we propose a new end-to-end low-precision workflow based on FlashAttention \cite{dao2024flashattention} to reduce the computation cost paid by softmax computation. Our workflow includes the introduction of low-precision arithmetic on the computation of softmax functions, and the reduction of the memory bandwidth required when transferring data between general-purpose and special-purpose accelerators by utilizing mixed-precision operations. The workflow can be expected to be integrated with quantized GEMMs and skip scheme like BLASST. To minimize the influence of new introduced features on the quality of large model inference, we apply the reorder of tile traversal and evaluating powers of two with integers. The rest of this paper is organized as follows. \S \ref{sec:related_work} reviews the related methodology to improve the throughput of computing attention scores. \S \ref{sec:approach} discusses the proposed methods of our workflow. \S \ref{sec:results} deploys the proposed methods to perform various typical tasks for AIGC, including reasoning and video generations from texts. Finally, we conclude with a summary of our contributions and results in \S \ref{sec:conclusion}.

%\subsection{Our results}
This paper introduces a novel low-precision workflow to address the softmax bottleneck in Transformer inference. By re-engineering the Flash-Attention \cite{dao2024flashattention} computation path, our method achieves two synergistic optimizations: (i) executing core softmax operations in low (8-bit) precision, and (ii) drastically cutting 50\% of the data movement bandwidth between general and special-purpose compute units via mixed-precision techniques. Our workflow is compatible with existing quantized GEMM backends and complementary to softmax-skipping approaches like BLASST \cite{yuan2025blasst}. Crucially, we maintain model accuracy via a key algorithmic innovation—block-aware precision rescaling—thereby overcoming the substantial accuracy degradation typical of direct low-precision attempts.

%\zy{The key idea is dynamically adjusting the precision for the local block by monitoring the difference between the block's local maximum and the global maximum accumulated from preceding blocks.} 
Our design adopts a default low-precision (FP8) path for forwarding GEMMs results. A safeguard mechanism conditionally triggers high-precision rescaling based on a dynamic error analysis. This analysis monitors the discrepancy between a block's local maximum and the global maximum propagated from previous blocks, dynamically signaling the need for increased precision. Empirically, rescaling and recalculation are required in only 5\% of cases in NLP and 10\% of cases in multi-modal tasks, preserving the bulk of the FP8 efficiency gains. This approach implements a signal-guided, dynamic precision adaptation strategy, which seamlessly toggles precision levels to optimize the accuracy-efficiency trade-off.

\subsection{Organization and overview}
The rest of this paper is organized as follows. Section \ref{sec:related_work} surveys prior art on accelerating attention. Section \ref{sec:approach} elaborates our proposed methodology. Section \ref{sec:results} evaluates our approach on diverse AIGC tasks, such as language and text-to-video generation. Section \ref{sec:conclusion} concludes the paper.

\section{Related Work}
\label{sec:related_work}

\paragraph{Low Precision and Sparse Attention}

A dominant trend in efficient transformer deployment is the aggressive quantization of linear operations, which reduces the bit-width of weights and activations to reduce memory bandwidth consumption and decrease the computational cost of GEMMs. The recent advancement in 4-bit training and inference \cite{nvidia2025pretraining} is one of the typical examples of this trend. However, a critical challenge emerges within the attention block: while the linear operations can be computed efficiently in low precision, standard softmax modules force related arithmetic to be carried out in high precision, causing a ``precision disparity''. 
%It is a precision disparity that the linear operations can be computated efficiently in low precision, while the softmax module is forced to high precision arithmetic. 
Consequently, the attention pipeline is forced to switch between low-precision GEMMs for linear operations and high-precision floating-point operations within softmax module \cite{zhang2025sageattention}, which costs computational resources spent on converting data between different precisions and wastes the extra numerical accuracy obtained by high-precision arithmetic.

% Rewrite:
% This precision disparity between components, along with the accompanied cost of converting between precisions and the remaining high-precision computations, has been the main bottleneck to adoping mixed-precision systems.
% The huge gap in the computing power required between the 4-bit GEMMS on specific-purpose accelerators and 32-bit operations on general-purpose accelerators, which is a difference of more than 2 orders of magnitudes in terms of floating operations per second (Flops) on the most advanced systems \cite{nvidiablackwellarichtecture}, results in low utilization rates of these specific-purpose accelerators.

The adoption of mixed-precision systems has been bottlenecked by precision disparity discussed above. The huge gap of computing power difference between the 4-bit GEMMs on specific-purpose accelerators and 32-bit operations on general-purpose accelerators, which is more than 2 orders of magnitudes difference in terms of floating operations per second (Flops) on the most advanced systems \cite{nvidiablackwellarichtecture}, results in low utilization rate of specific-purpose accelerators. The utilization rate is only around 30\% \cite{zhang2025sageattention} on consumer-level devices, and expected to be even lower on more advanced systems. Although recently proposed sparse schemes, e.g., BLASST \cite{yuan2025blasst}, can skip 50\% of the computation in attention modules without loss in accuracy. 
Yet, for the remaining, unavoidable computations that must be executed, the high-precision softmax operation continues to create a disparity with the low-precision GEMMs. This persistent precision mismatch leaves the hardware utilization bottleneck largely unaddressed, underscoring the need for fundamental low-precision softmax designs.
%, the marginal gain is still very little without the development of high-throughput softmax modules in case when the proposed scheme is integrated with low-precision quantized GEMMs.

\paragraph{Co-designed Approximations for Softmax}

On top of algorithmic innovations like Flash-Attention \cite{dao2024flashattention} that aim to reduce the memory footprint of attention mechanisms, another parallel line of research focuses on the softmax operation itself. These works employ hardware-software co-design to accelerate or approximate the expensive exponentiation and normalization steps inherent to the softmax operation.

One prominent strategy, exemplified by Stevens et al. \cite{Stevens2021Softermax}, seeks to approximate the softmax operation to reduce its hardware cost. Their work, ``Softermax'', proposes replacing high-precision floating-point exponentiation with a low-cost, piecewise first-order approximation of the $\mathtt{exp2}$ function using integer arithmetic. While this approach achieves notable gains in area and energy efficiency, it also introduces fundamental limitations that likely hindered its widespread adoption: the first-order approximation can incur non-negligible numerical errors, which poses a potential risk for tasks involving long sequences or requiring high numerical fidelity because the error can accumulate across iterations. Furthermore, the proposed design focuses on the computation itself but does not optimize the memory bandwidth at the \textit{input} of the softmax module. The preceding attention scores are typically still calculated and transferred in high precision, meaning the memory bandwidth preceding the softmax module remains a performance and energy constraint, limiting the end-to-end benefit.

The attention module has multiple computation operations that can be fused together to minimize intermediate data read and write operations and enable low-precision approximations to reduce the intermediate memory access expense. Alexandridis et al. \cite{alexandridis2025low} propose a solution that fuses exponentiation and multiplication into a single hardware unit which adopts shift operations to save the computation delay and avoids the computation of exponents in high precision and uses shift operations to save the computation delay. Similarly, Wang et al. \cite{wang2025sole} take a broader co-design view, not only fusing internal softmax steps but also combining softmax and the layer normalization operations into a unified hardware pipeline. This fuses two consecutive stages in attention and greatly reduces the control and data transfer overhead across multiple layers. Although these fusion-based methods align with the memory-bound nature of attention, their shared major drawback is increased hardware specialization and inflexibility for supporting emerging algorithmic variants or alternative normalization schemes.

\paragraph{Low Precision Data Formats}

Several data formats exist for low-precision floating numbers. For 16-bit representations, FP16 and BF16 are the most common ones. BF16 retains the same number of exponent bits as FP32, resulting in a larger dynamic range compared to FP16. This makes BF16 more robust against outliers and better at capturing gradients during training. Thus, BF16 is preferred for training neural networks. When it comes to 8-bit floating numbers, a wider variety of formats are available. Single numbers can typically be represented using E4M3 (4 exponent bits and 3 mantissa bits) or E5M2 (5 exponent bits and 2 mantissa bits). To balance memory footprint with numerical accuracy, MXFP8 \cite{mishra2025recipe} has been recently introduced for training LLMs. In MXFP8, multiple numbers share a common scaling factor while each is stored in E4M3 format. After scaling, MXFP8 can achieve a dynamic range comparable to that of E5M2 but with higher accuracy. MXFP8 is especially suitable for linear operations such as matrix multiplications, where scaling factors and data can be efficiently separated. An alternative approach to extend the dynamic range of data is to consider the recently introduced 8-bit format - HiF8 \cite{luo2024ascend}. HiF8 employs a dynamic exponent-mantissa coding scheme that simultaneously expands the representable range and maintains accuracy. Unlike MXFP8, it does not rely on shared scaling factors, making it a suitable choice for complex nonlinear computations.

\newcommand{\rowmax}{\mathtt{rowmax}}
\newcommand{\diag}{\mathtt{diag}}

\section{Approach}
\label{sec:approach}

Consider query $\mathbf{Q} = [\mathbf{Q}_0; \mathbf{Q}_1; \dots; \mathbf{Q}_{N_Q}] \in \mathbb{R}^{s_q \times d}$, key $\mathbf{K} = [\mathbf{K}_0; \mathbf{K}_1; \dots; \mathbf{K}_{N_K}] \in \mathbb{R}^{s_k \times d}$ and value $\mathbf{V} = [\mathbf{V}_0; \mathbf{V}_1; \dots; \mathbf{V}_{N_K}] \in \mathbb{R}^{s_k \times d}$, where $s_q$ and $s_k$ is the sequence length of query and key, and $d$ is the head dimension. The standard attention is computed as follows:
\begin{equation}
    \mathbf{O} = \mathbf{P} \mathbf{V} = \mathtt{softmax} (\mathbf{S}) \mathbf{V} = \mathtt{softmax} \left( \dfrac{\mathbf{Q} \mathbf{K}^\intercal}{\sqrt{d}} \right) \mathbf{V} ,
    \label{eq:attention}
\end{equation}
where $\mathbf{S}={(\mathbf{Q} \mathbf{K}^\intercal)}/{\sqrt{d}}$ is the scaled dot-product scores. The core component of this computation is the nonlinear softmax function \cite{bishop2006pattern}, which normalizes the scaled dot-product scores to produce a valid probability distribution over keys for each query.

As $s_q$ and $s_k$ can be very large in the general purpose tasks, Flash Attention (FA) \cite{shah2024flashattention, dao2024flashattention} is employed to accelerate computation and avoids storing large-sized intermediate results, most notably the attention score matrix $\mathbf{S} \in \mathbb{R}^{s_q \times s_k}$. The core iterative step of FA can be expressed as follows:
\begin{equation}
    \begin{aligned}
        \mathbf{S}_i^{(j)} = & \mathbf{Q}_i \mathbf{K}_j^\intercal \cdot \frac{\log_2 e}{\sqrt{d}} \\
        \mathbf{m}^{(j)}_i = & \rowmax ([\mathbf{m}^{(j - 1)}_i,\mathtt{rowmax} (\mathbf{S}^{(j)}_i)]) \\
        \mathbf{P}_i^{(j)} = & 2^{\mathbf{S}^{(j)}_i - \mathtt{rowfill}(\mathbf{m}^{(j)}_i)} \\
        \mathbf{d}^{(j)}_i = & \diag (2^{\mathbf{m}^{(j - 1)}_i - \mathbf{m}^{(j)}_i}) \mathbf{d}^{(j - 1)}_i + \mathbf{P}_i^{(j)}\mathbf{1} \\
        \widetilde{\mathbf{O}}^{(j)}_i = & \mathtt{diag} (2^{\mathbf{m}^{(j - 1)}_i - \mathbf{m}^{(j)}_i}) \widetilde{\mathbf{O}}^{(j - 1)}_i + \mathbf{P}_i^{(j)} \mathbf{V}_j \;,
    \end{aligned}
    \label{eq:flash-attention}
\end{equation}
in which $\mathbf{S}_i^{(j)}$ denotes the scaled dot-product scores between the 
$i$-th query tile and 
the $j$-th key tile, $\mathbf{m}_i^{(j)}$ denotes the vector storing the row-wise maximum values used for numerical stabilization, $\mathbf{d}_i^{(j)}$ represents the denominator vector in the online softmax computation, and $\widetilde{\mathbf{O}}_i^{(j)}$ denotes the corresponding un-normalized attention output. Moreover, $\mathtt{rowfill}(\cdot)$ expands a vector into a matrix by replicating each element of the input vector across an entire row, such that every row of the resulting matrix is a constant vector equal to the corresponding element of the input. Conversely, $\mathtt{rowsum}(\cdot)$ reduces a matrix to a vector by summing the elements along each row.

In most of the open-source models \cite{wan2025wan}, the query, key and value matrices ($\mathbf{Q}, \mathbf{K}, \mathbf{V}$) are typically stored as BF16 format. To reduce the memory bandwidth requirements and computational resources, post-training quantization (PTQ)~\cite{Frantar2022GPTQAP} or activation-aware weight quantization (AWQ)~\cite{Lin2023AWQAW} can be applied, typically with negligible degradation in model accuracy.
 Micro-scaling \citep[e.g.,][]{zhang2025sageattention} can alternatively reduce the cost of attention computation as well. 
 %These techniques mainly consider low precision representation of $\mathbf{Q}$, $\mathbf{K}$ and $\mathbf{V}$, e.g., FP8 or FP4, but leave the intermediate results, e.g., $\mathbf{S}$ and $\mathbf{P}$ at higher precisions. 
 Consequently, while existing methods successfully reduce the precision of $\mathbf{Q}$, $\mathbf{K}$, and $\mathbf{V}$ to as low as FP8 or even FP4, the precision of $\mathbf{S}$
and the subsequent softmax normalization $\mathbf{P}$ remain constrained to be as high as BF16/FP16 or even FP32. This leaves the precision disparity within the attention block unresolved.

Meanwhile, modern accelerators mitigate the workload of matrix-matrix products by employing special-purpose accelerators, e.g., Tensor cores on Nvidia GPUs. It splits the procedure into multiple individual stages running on special-purpose and general-purpose accelerators separately. This results in potentially varying precision requirements between different stages. In the case of Flash Attention (\ref{eq:flash-attention}),  although the accumulator produces outputs in 32-bit precision, maintaining the same precision throughout the softmax calculation is unnecessary. By leveraging quantization techniques, the matrix product $\mathbf{P} \mathbf{V}$ in \Cref{eq:attention} can even take 4-bit numbers as the inputs without much loss. This suggests that performing all softmax operations in 32-bit precision introduces significant redundancy.

%This observation motivates us to explore computing intermediate numbers in lower precisions with only a minor degradation of end-to-end model accuracy. In the remainder of this section, we discuss several approaches to save the computational resources for attention calculation in \Cref{eq:attention}. As $\mathbf{P}$ only requires a low-precision representation, we first propose using 8-bit precision formats as the input and output data format of exponent operations in Section \ref{subsec:exp_B8} to compute $\mathbf{P}$ directly without quantization. \qgl{Next, we discuss about saving scores $\mathbf{S}$ to 8-bit numbers to reduce the memory bound when transferring data from special-purpose accelerators to general-purpose accelerators in Section \ref{subsec:scores_B8}. Moreover, we consider using rounded floating numbers to improve the accuracy of exponent operations in Section \ref{subsec:int_exp}. Finally, we propose using HiF8 \cite{luo2024ascend} to simplify the implementation of discussed methods and summarize the entire workflow in Section \ref{subsec:summary}.}

Motivated by this observation, we investigate efficient attention computation using reduced numerical precision, which can substantially save resources while preserving end-to-end model accuracy. This section details our strategies to reduce the cost of computing attention in \Cref{eq:attention}. Our key insight is that the matrix $\mathbf{P}$ inherently tolerates low precision. Consequently, we first propose computing $\mathbf{P}$ directly in 8-bit precision for the exponential operations without quantization steps (Section\ref{subsec:exp_B8}). To further reduce memory traffic bottleneck when shuttling data between special-purpose and  general-purpose accelerators, we propose storing the attention scores $\mathbf{S}$ in 8-bit format (Section \ref{subsec:scores_B8}). 
%We also consider using rounded floating to improve the numerical accuracy of these low-precision exponentials (Section \ref{subsec:int_exp}). 
To combat the multiplication error accumulation in online softmax, we further propose to compute exponentials using base-2 raised to integer exponents. This approach significantly improves numerical accuracy while maintaining low-precision efficiency (Section \ref{subsec:int_exp}).
All these methods are integrated and simplified using the HiF8 format \cite{luo2024ascend}, with the complete workflow summarized in Section \ref{subsec:summary}.

\subsection{8-bit Exponents ($\mathbf{P}$)}
\label{subsec:exp_B8}

As noted above, the exponentiation operation does not require high precision. For example, in most accelerated inference scenarios \cite{zhang2025sageattention}, the exponentiated result $\mathbf{P}$ can be cast to lower precision before the matrix multiplication with $\mathbf{V}$. Since $\mathbf{S}_i^{(j)} - \mathtt{rowfill}(\mathbf{m}_i^{(j)})$ is predominantly negative and confined to a moderate numerical range
%provided a proper head dimension and normalization scale factor are chosen, we consider utilizing 8-bit low-precision formats as its representation piror to computing powers of two. 
(given appropriate head dimension and normalization scaling), we represent it using an 8-bit low-precision format prior to exponentiation.

\begin{figure}[htpb]
    \centering
    \begin{subfigure}{0.48\linewidth}
        \centering
        \includegraphics[width=\linewidth]{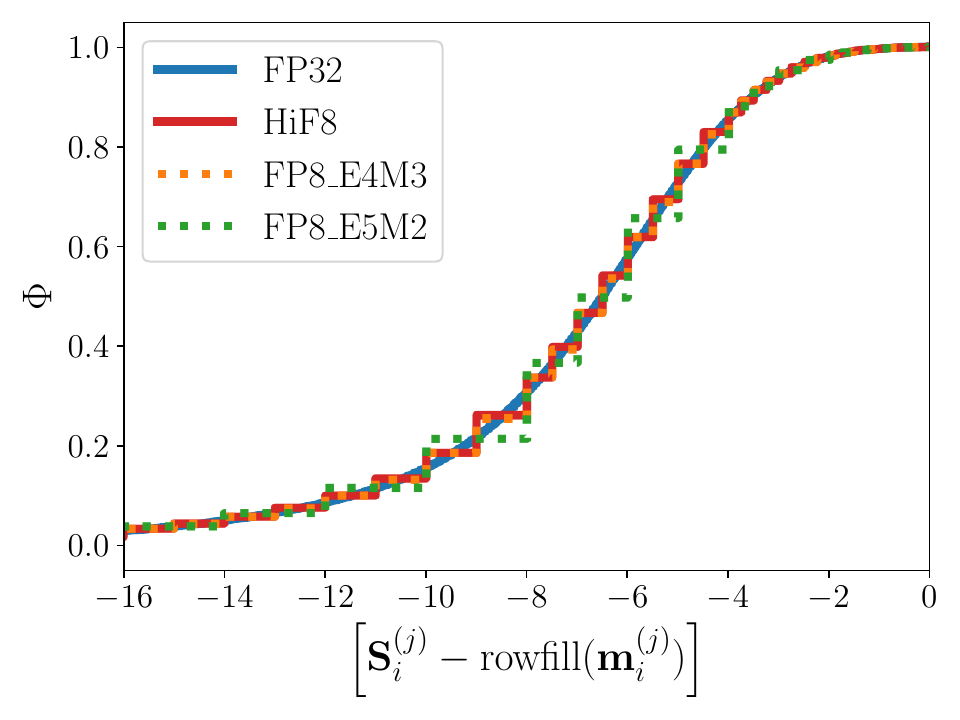}
        \caption{The cumulative distribution function $\Phi$ of the difference between the scores and the row maximums $\mathbf{S}_i^{(j)} - \mathrm{rowfill}(\mathbf{m}_i^{(j)})$.}
    \label{fig:scores_distribution}
    \end{subfigure}
    \hfill
    \begin{subfigure}{0.48\linewidth}
        \centering
        \includegraphics[width=\linewidth]{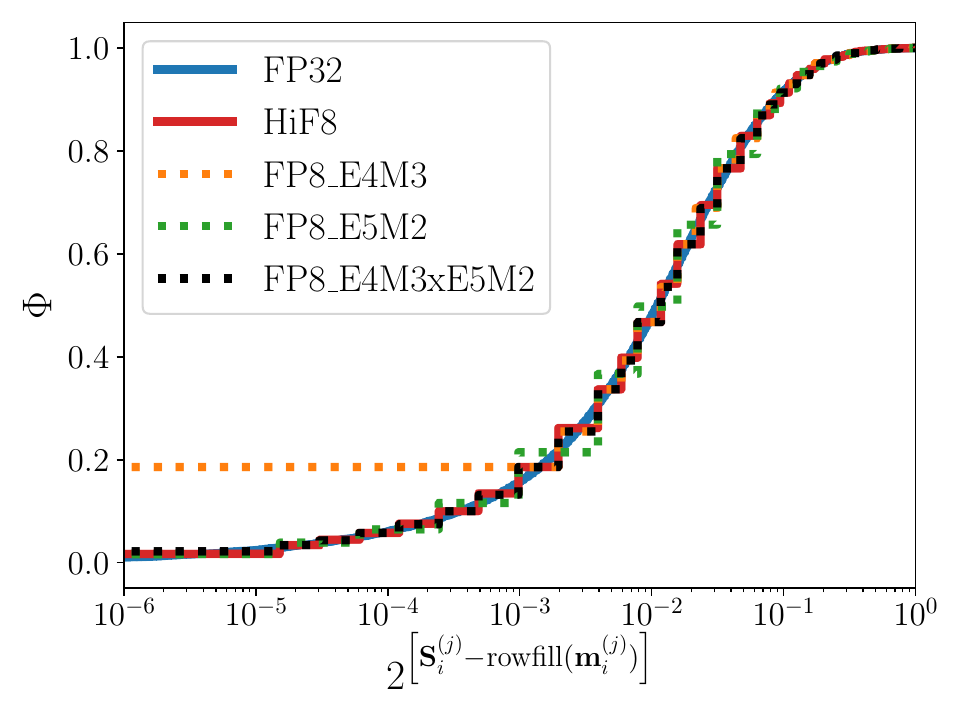}
        \caption{The cumulative distribution function $\Phi$ of the two to the difference between the scores and the row maximums $2^{\mathbf{S}_i^{(j)} - \mathrm{rowfill}(\mathbf{m}_i^{(j)})}$.}
    \label{fig:power_distribution}
    \end{subfigure}
    \caption{The cumulative distribution function $\Phi$ of different data formats. The scores are dumped from Qwen3-30B-A3B-Instruct model.}
    \label{fig:distribution_analysis}
\end{figure}

There are two typical FP8 formats-E4M3 and E5M2, where E4M3 is strongly preferred for representing $\mathbf{S}_i^{(j)} - \mathtt{rowfill}(\mathbf{m}_i^{(j)})$ due to its larger mantissa. This yields a quantized distribution that more closely approximates the original FP32 values, as shown in \Cref{fig:scores_distribution}.
Conversely, exponentiating two by this difference compresses the range of $\mathbf{P}_i^{(j)}$ within $[0,1]$( see \Cref{fig:power_distribution}), presenting a distinct challenge: for values in $[0, 2^{-9})$, E4M3 lacks sufficient resolution, which limits the model's expressive capacity. E5M2, with its larger dynamic range, is therefore better suited for representing $\mathbf{P}$.

These observations motivates a ``mixed format'' strategy, ranther than using only E4M3 or E5M2 exclusively. As shown in \Cref{fig:scores_distribution}, the distributions of the powers-of-two operation results under different input/output precision formats are compared against the FP32 baseline. 

%The orange and green curves represent the pure E4M3 and pure E5M2 configurations, respectively, while the red curve corresponds to a mixed-format approach—using E4M3 for the input and E5M2 for the output. This mixed format yields the closest approximation to the FP32 reference distribution. However, it introduces additional complexity in both software and hardware implementation. Consequently, as discussed in \S\ref{subsec:summary}, adopting HiF8 offers a more practical and efficient alternative.
The orange and green curves represent pure E4M3 and pure E5M2 configurations, respectively, while the red curve corresponds to the mixed-format approach (E4M3 input, E5M2 output). This hybrid scheme most closely approximates the FP32 reference distribution. However, it introduces added complexity in both software and hardware implementation. Consequently, as discussed in Section \ref{subsec:summary}, we adopt HiF8 \cite{luo2024ascend} as a more practical and efficient alternative.

\subsection{Towards 8-bit Attention Scores ($\mathbf{S}$)}
\label{subsec:scores_B8}
Our computational reorganization is motivated by a hardware-aware precision trade-off: although the score matrix  $\mathbf{S}_i^{(j)}$ itself lacks sufficient precision in 8-bit representation, the difference form $\mathbf{S}_i^{(j)}-\mathtt{rowfill}(\mathbf{m}_i^{(j)})$ can be cast to 8-bit with acceptable accuracy. In the standard Flash Attention pipeline, however, this subtraction occurs on the general-purpose accelerators (e.g., Vector/Cuda Cores) after $\mathbf{S}_i^{(j)}$ is transferred from the special-purpose accelerators (e.g., Cube/Tensor Cores). Therefore, simply casting the subtraction result %$\mathbf{S}_i^{(j)}-\mathtt{rowfill}(\mathbf{m}_i^{(j)})$ 
to 8-bit post-computation does not reduce the dominant memory traffic of $\mathbf{S}_i^{(j)}$. To capture this benefit, the low-precision computation casting must be moved earlier in the pipeline.

We therefore propose to offload the critical subtraction step. Instead of transferring the full-precision $\mathbf{S}_i^{(j)}$
as in \Cref{eq:flash-attention}, we let special-purpose accelerators compute:
\begin{equation}
\mathbf{T}_i^{(j)} = \mathbf{S}_i^{(j)} - \mathtt{rowfill}(\widetilde{\mathbf{m}}_i^{(j)}) \;, 
\end{equation}
where $\widetilde{\mathbf{m}}_i^{(j)}$ is an ``approximate'' row-wise maximum, initialized as $\widetilde{\mathbf{m}}_i^{(j)} = \mathbf{m}_i^{(j-1)}$ (i.e., the maximum from the previous iteration). $\mathbf{T}_i^{(j)}$ is then transferred to general-purpose accelerators in 8-bit low precision form. An update of $\widetilde{\mathbf{m}}_i^{(j)}$ and a recalculation of $\mathbf{T}_i^{(j)}$ could be triggered conditionally, based on the value range of $\mathbf{T}_i^{(j)}$ and controlled by a restart threshold $\lambda$.

Once the ``approximate'' row-wise maximums $\widetilde{\mathbf{m}}_i^{(j)}$ are obtained, they are transferred to the special-purpose accelerators. The subtraction is then performed before the result is sent to the general-purpose accelerators, effectively operating on lower-precision data earlier. 
This approximation is effective due to the systematic bias in the distribution of row-wise maxima in attention scores \cite{kaul2024attention}, where certain tiles are significantly more likely to contain the global maximum. For example, in auto-regressive language modeling, the first and last tiles often exhibit this property.

We denote the set of such critical tiles as $\mathcal{B}_0$, and the remaining tiles as $\mathcal{B}_1$. For tiles in  $\mathcal{B}_0$, which are evaluated first,  $\mathbf{m}_i^{(j)}$ is already a close approximation to the final global maximum. Consequently, for a majority of subsequent tiles,  ${\mathbf{T}}_i^{(j)}$ (calculated using the approximate maximum) is accurate enough for downstream computations, and preserving ${\mathbf{T}}_i^{(j)}$ in high precision yields diminishing returns.

\subsection{Reduce Accumulated Error of Power-of-2 Computations}
\label{subsec:int_exp}

\begin{figure}[htbp]
    \centering
    \includegraphics[width=0.75\linewidth]{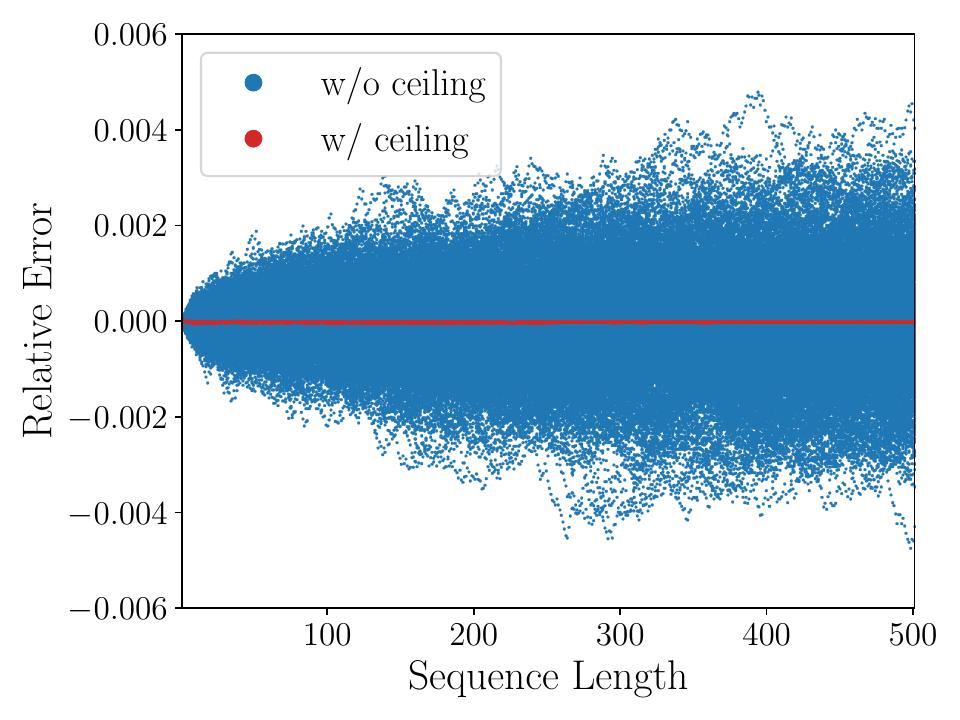}
    \caption{Illustration of numerical error in accumulative products: a comparison between exponents of 2 with integers and direct exponents of 2 against FP32 ground truth.}
    \label{fig:int_exp2}
\end{figure}

%Similar to \cite{Stevens2021Softermax}, we also compute powers of 2 with integers but it is limited to the rescaling factor when updating $\mathbf{d}_i^{(j-1)}$ and $\widetilde{\mathbf{O}}_i^{(j-1)}$. As these updates are performed multiple times as the scaling of the length of the sequence, the numerical error generated from low precision operations can be accumulated across iterations. A simple numerical experiment, illustrated in \Cref{fig:int_exp2}, demonstrates that the numerical error can propagated as the sequence length increases. Thus, we propose using the integer powers of two to reduce the accumulated multiplicative error. In contrast with normal exponent operations or power of two with arbitrary numbers, the powers of two with integers would not introduce any numerical error. As seen in \Cref{fig:int_exp2}, the numerical error can be easily bounded when using the powers of two with integers to update the cumulative operations. Therefore, the update of $\mathbf{m}_i^{(j)}$ in \Cref{eq:flash-attention} requires rounding of each entry in $\mathbf{S}_i^{(j)}$. The reorganized formulation of attention calculation is summarized in \Cref{eq:proposed_attention}.
Following \cite{Stevens2021Softermax}, we restrict the use of powers of two to integer exponents, specifically for the rescaling factor when updating $\mathbf{d}_i^{(j)}$ and $\widetilde{\mathbf{O}}_i^{(j)}$. 
Because these updates recur as the sequence length scales, low-precision operations can cause numerical errors to accumulate over iterations. \Cref{fig:int_exp2} illustrates how this error propagates with increasing sequence length. To mitigate the accumulated multiplicative error, we propose using integer powers of two. Crucially, unlike general exponentiation or powers of two with arbitrary (non-integer) exponents, integer powers of two are exact and introduce no additional numerical error. Consequently, as evidenced in \Cref{fig:int_exp2}, the error remains tightly bounded when integer powers of two are used for cumulative updates. This approach necessitates rounding each entry in $\mathbf{S}_i^{(j)}$ to update $\mathbf{m}_i^{(j)}$ in \Cref{eq:flash-attention}. The complete reorganized attention formulation is presented in \Cref{algo:workflow}.

\subsection{Summary of the Proposed Method}
\label{subsec:summary}
The complete workflow is summarized in \Cref{algo:workflow}. Our method modifies the standard Flash Attention (\Cref{eq:flash-attention}) by  by integrating the subtraction of the row maximum from the scores directly into the computation of the $\mathbf{Q}_i\mathbf{K}_j^\intercal$ products. 

\begin{algorithm}[htbp]
    \caption{The Proposed Workflow.}
    \label{algo:workflow}
    \begin{algorithmic}[1]
        \STATE {\bfseries Input:} Query $\mathbf{Q}_i$, key $\mathbf{K}$ and value $\mathbf{V}$. \STATE {\bfseries Parameter:} Restart threshold $\lambda$.
        \FOR {$j \in \mathcal{B}_0$}
        \STATE $\mathbf{S}_i^{(j)}=\mathbf{Q}_i \mathbf{K}_j^\intercal \cdot \frac{\log_2 e}{\sqrt{d}}$, in high precision
        \STATE $\mathbf{m}_i^{(j)}= \mathtt{max}(\mathbf{m}_i^{(j-1)},{\color{red}\lceil} \mathtt{rowmax} (\mathbf{S}^{(j)}_i) {\color{red} \rceil})$
        \STATE $\mathbf{P}_i^{(j)} = 2^{\mathbf{S}^{(j)}_i - \mathtt{rowfill}({\mathbf{m}}^{(j)}_i})$
        \STATE $\mathbf{d}_i^{(j)} = \diag (2^{\mathbf{m}^{(j - 1)}_i - \mathbf{m}^{(j)}_i}) \mathbf{d}^{(j - 1)}_i + \mathbf{P}_i^{(j)}\mathbf{1}$ 
        \STATE $\widetilde{\mathbf{O}}_i^{(j)} = \mathtt{diag} (2^{\mathbf{m}^{(j - 1)}_i - \mathbf{m}^{(j)}_i}) \widetilde{\mathbf{O}}^{(j - 1)}_i + \mathbf{P}_i^{(j)} \mathbf{V}_j  $ 
        \ENDFOR
        \FOR {$j \in \mathcal{B}_1$}
        \STATE $\widetilde{\mathbf{m}}_i^{(j)}= \mathbf{m}_i^{(j-1)}$
        \STATE $\mathbf{T}_i^{(j)}=\mathbf{Q}_i \mathbf{K}_j^\intercal \cdot \frac{\log_2 e}{\sqrt{d}} - \mathtt{rowfill}(\widetilde{\mathbf{m}}^{(j)}_i)$, \textcolor{red}{in 8-bit}
        \IF {$\lceil \mathtt{rowmax} (\mathbf{T}^{(j)}_i)  \rceil> \lambda$}   
            \STATE $\mathbf{S}_i^{(j)}=\mathbf{Q}_i \mathbf{K}_j^\intercal \cdot \frac{\log_2 e}{\sqrt{d}}$, in high precision
            \STATE $\mathbf{m}_i^{(j)}= \mathtt{max}(\mathbf{m}_i^{(j-1)},{\color{red}\lceil} \mathtt{rowmax} (\mathbf{S}^{(j)}_i) {\color{red} \rceil})$
            \STATE $\widetilde{\mathbf{m}}_i^{(j)}= \mathbf{m}_i^{(j)}$
            \STATE $\mathbf{T}_i^{(j)}=\mathbf{S}_i^{(j)} - \mathtt{rowfill}(\widetilde{\mathbf{m}}^{(j)}_i)$, \textcolor{red}{in 8-bit}
        \ELSE
         \STATE $\mathbf{m}_i^{(j)}= \mathbf{m}_i^{(j-1)} + \mathtt{max}(0, {\color{red}\lceil} \mathtt{rowmax} (\mathbf{T}^{(j)}_i)  {\color{red}\rceil})$
        \ENDIF
        \STATE $\mathbf{P}_i^{(j)} = 2^{\mathbf{T}^{(j)}_i}$, \textcolor{red}{in 8-bit}
        \STATE $\mathbf{d}_i^{(j)} = \diag (2^{\mathbf{m}^{(j - 1)}_i - \mathbf{m}^{(j)}_i}) \mathbf{d}^{(j - 1)}_i + \mathbf{P}_i^{(j)}\mathbf{1}$ 
        \STATE $\widetilde{\mathbf{O}}_i^{(j)} = \mathtt{diag} (2^{\mathbf{m}^{(j - 1)}_i - \mathbf{m}^{(j)}_i}) \widetilde{\mathbf{O}}^{(j - 1)}_i + \mathbf{P}_i^{(j)} \mathbf{V}_j  $ 
        \ENDFOR
        \STATE {Normalization: $\mathbf{O}_i = \mathtt{diag}^{-1}(\mathbf{d}_i) \widetilde{\mathbf{O}}_i$.}
        \STATE {\bfseries Return:} Normalized attention score $\mathbf{O}_i$.
    \end{algorithmic}
\end{algorithm}
This can be fused with the dequantization step when $\mathbf{Q}_i$ and $\mathbf{K}_j$ are hardware-quantized. The row max $\mathbf{m}_i^{(j)}$ is always rounded to the nearest integer for enhancing the numerical accuracy when updating $\mathbf{d}_i^{(j)}$ and $\widetilde{\mathbf{O}}_i^{(j)}$. 
Furthermore, leveraging the prior pattern of the row-wise maximum discussed in Section \ref{subsec:scores_B8}, the value of $\mathbf{m}_i^{(j)}$ can often be predicted with high confidence by examining a small subset of tiles in advance. This allows the update for $\mathbf{m}_i^{(j)}$ to be skipped in the majority of tile iterations, improving efficiency. For hardware optimization, the powers of 2 in the calculation of $\mathbf{P}_i^{(j)}$ are computed using the HiF8 format, which reduce circuit area and increases throughput for this nonlinear operation.

% \begin{equation}
%     \small
%     \begin{aligned}
%         \widetilde{\mathbf{S}}_i^{(j)} = & \mathbf{Q}_i \mathbf{K}_j^\intercal \cdot \frac{\log_2 e}{\sqrt{d}} - \mathbf{m}_i^{(j-1)} \\
%         \widetilde{\mathbf{m}}^{(j)}_i = &
%         \mathtt{rowmax}([\mathbf{0}, \lceil \mathtt{rowmax} (\widetilde{\mathbf{S}}^{(j)}_i) \rfloor ]) \\
%         \mathbf{m}_i^{(j)} = & \mathbf{m}_i^{(j-1)} + \widetilde{\mathbf{m}}^{(j)}_i \\
%         \mathbf{P}_i^{(j)} = & 2^{\widetilde{\mathbf{S}}^{(j)}_i - \widetilde{\mathbf{m}}^{(j)}_i} \\
%         \mathbf{d}^{(j)}_i = & \mathtt{diag} (2^{-\widetilde{\mathbf{m}}^{(j)}_i}) \mathbf{d}^{(j - 1)}_i + \mathtt{rowsum}( \mathbf{P}_i^{(j)} ) \\
%         \widetilde{\mathbf{O}}^{(j)}_i = & \mathtt{diag} (2^{-\widetilde{\mathbf{m}}^{(j)}_i}) \widetilde{\mathbf{O}}^{(j - 1)}_i + \mathbf{P}_i^{(j)} \mathbf{V}_j \;.
%     \end{aligned}
% \end{equation}

To control the accuracy of the attention outputs, the accuracy of approximate row-wise maximum $\widetilde{\mathbf{m}}_i^{(j)}$ can be adjusted dynamically based on the value of $\mathtt{rowmax}(\mathbf{T}_i^{(j)})$. Specifically, if the maximum entry in $\mathtt{rowmax}(\mathbf{T}_i^{(j)})$ exceeds a predefined threshold $\lambda$, the scores are recalculated and stored in high precision for subsequent processing on general-purpose accelerators. The numerical results in \Cref{tab:restart-analysis} indicate that such rescaling and recalculation events are very rare, as $\mathtt{rowmax}(\mathbf{T}_i^{(j)})$ is typically non-positive. Consequently, with high probability, the recalculation of $\mathbf{T}_i^{(j)}$, as well as the rescaling of $\mathbf{d}_i^{(j-1)}$ and $\widetilde{\mathbf{O}}_i^{(j-1)}$ can be avoided.

\section{Ablation Experiments}
\label{sec:results}
In this section, we examine the proposed workflow for both language model and multi-modal model.
\subsection{Experimental Setups}
We evaluate our proposed method under three configurations:
\begin{itemize}
    \item \textbf{Default}: In this configuration, We compute the $\mathtt{softmax}(\cdot)$ as stated in~\Cref{eq:flash-attention} with inputs stored in BF16 and all operations within the softmax  performed in FP32.
    \item \textbf{Exp2 HiF8}: This configuration is identical to the \textbf{Default} configuration, except that all power-of-two exponentiation operations in the softmax use the HiF8 data format. This targets the most computationally expensive operation in the softmax by employing HiF8 arithmetic.%time consuming operation in softmax module by HiF8 arithmetic.
    \item \textbf{E2E HiF8}: This configuration computes the $\mathtt{softmax}$ as defined in~\Cref{algo:workflow}, with inputs stored in BF16 and HiF8 extended to nearly all operations within the softmax module. The first tile in each input sequence is assigned to $\mathcal{B}_0$, and the remaining tiles to $\mathcal{B}_1$, as specified in~\Cref{algo:workflow}. Unless otherwise specified, we set $\lambda=1$.
     % The most prominent feature of this setup is saving scores to HiF8 when passing from special-purpose accelerators to general-purpose accelerators as highlighted in \S \ref{subsec:scores_B8}. \qgl{Since Qwen3 is a causal language model, the first and last tiles in each input sequence are calculated with high precision, while the remaining tiles use low-precision arithmetic, as discussed in \S \ref{subsec:scores_B8}.} Moreover, \qgl{to minimize the possibility of restart(rewrite this sentence)}, $\lambda$ is set to 4 when recomputing and re-passing $\widetilde{\mathbf{S}}_i^{(j)}$ in high precision.
\end{itemize}
%All experiments in these configurations use the same random seed to ensure consistent randomness across configurations, thereby guaranteeing the fairness of the ablation study.
All experiments use the same random seed to ensure consistent randomness across configurations, guaranteeing a fair ablation study.
\subsection{The Language Models' Accuracy}

Since large-scale models are more robust to numerical errors from low-precision arithmetic, small-scale models offer a more challenging and effective test for validating our proposed method.
To evaluate the sensitivity of low-precision operations in {natural language processing}, we focus on small-scale models containing fewer than 50B parameters—approximately one-tenth the size of flagship open-source models such as DeepSeek-V3\cite{deepseekai2025deepseekv3}. 
We primally focus on Qwen3-30B-A3B-Instruct-2507~\cite{qwen3technicalreport} and also adopt the smaller Llama-3.1-8B~\cite{grattafiori2024llama3} as the language models in various benchmarks.

In order to verify the advantage of HiF8 over other FP8 formats, we first use the common word extraction task (CWE) in RULER~\cite{hsieh2024ruler}, the empirically sensitive sub-task in RULER when varying the input sequence lengths, for the comparison. We consider three different sequence lengths of input tokens, 32K, 64K and 96K, in the task. For each sequence length, the test dataset is composed by 100 randomly generated samples. Three additional setups are considered by following the setup of \textbf{Exp2 HiF8} except with different input/output data formats for exponentiation:
\begin{itemize}
\item[1)] \textbf{Exp2 E4M3:} using E4M3 for both input/output,
\item[2)] \textbf{Exp2 E5M2:} using E5M2 for both input/output, 
\item[3)] \textbf{Exp2 E4M3xE5M2:} using E4M3 for input and E5M2 for output,
\item[4)] \textbf{Naive E2E:} directly converting scores to HiF8 before subtraction with row maximum.
\end{itemize}
As shown in \Cref{tab:cwe_qwen3}, the \textbf{Exp2 E4M3} format incurs substantial accuracy loss on the CWE task, aligning with the power distribution analysis in \Cref{fig:power_distribution} and the discussion in \Cref{subsec:exp_B8}. A detailed comparison at sequence length 64K (where the \textbf{Default} baseline exceeds 60\% accuracy) reveals that \textbf{Exp2 E4M3xE5M2} and our proposed \textbf{Exp2 HiF8} consistently outperform other FP8 configurations.  This validates the advantage of the HiF8 format for low-precision softmax exponentiation, a design choice justified by its adaptive range and implementation simplicity. Converting to FP8 after subtraction and the restart mechanism is essential to the success of \textbf{E2E HiF8} setup for ensuring the models' accuracy when comparing the results from \textbf{Naive E2E} which directly converting scores to HiF8 without modifying the original workflow of Flash-Attention.
%Consistent with the analysis shown in~\Cref{fig:power_distribution} and the discussion in~\Cref{subsec:exp_B8},  shows an obvious degradation in CWE task, especially when it gets long input sequences. By comparing the results with the sequence length of 64K, as it has an accuracy more than 60\% in \textbf{Default} setup which can provide a better insight, \textbf{Exp2 E4M3xE5M2} and \textbf{Exp2 HiF8} show an advantage over the other two setups using FP8. As pointed in~\Cref{subsec:exp_B8}, this simple test helps demonstrate the superiority of adopting HiF8 for exponentiation in softmax, especially when considering its adaptivity and simplicity in both software and hardware implementation.}

The evaluation of the proposed method on language reasoning is conducted on the MMLU-Pro benchmark suite~\cite{wang2024mmlupro}, GPQA-Diamond~\cite{rein2023gpqa}, and MATH-500 dataset~\cite{lightman2023lets}.
The results are reported in \Cref{tab:scores_models}, only using a single shot for each test sample.

Comparing to \textbf{Default} setup, both \textbf{Exp2 HiF8} and \textbf{E2E HiF8} show a comparable overall accuracy with at most 1\% degradation in the quantitative results across different models and different benchmark suites, except GPQA-Diamond on Llama-3.1.
This is a highly promising result, as adopting HiF8 can significantly reduce exponentiation operations consumption and thereby improve the computational throughput of the softmax module. Moreover, \textbf{E2E HiF8} offers even greater throughput gains by at least halving the memory bandwidth required when transferring data from special-purpose accelerators to general-purpose ones.

However, certain sub-tasks exhibits notable accuracy fluctuation which are all highlighted in red in \Cref{tab:mmlu-qwen3}-\ref{tab:math500-llama}. 
This suggests that different tasks vary considerably in their sensitivity to numerical precision. 
To mitigate this issue and ensure robust performance across diverse downstream tasks, post-processing, e.g., PTQ, is required to make the model aware of low-precision arithmetic and improve the accuracy for different kinds of downstream tasks before deploying the proposed inference acceleration scheme.

\begin{table}[htpb]
    \caption{Accuracy of Qwen3-30B-A3B on CWE with various input lengths of tokens. The accuracy is calculated in percentage and higher values are preferable.}
    \label{tab:cwe_qwen3}
    \vskip 0.15in
    \begin{center}
        \begin{tabular}{lcccc}
            \toprule
            \diagbox{Setup}{Seq Len} & \textbf{32K} & \textbf{64K} & \textbf{96K} \\
            \midrule
            \textbf{Default} & 88.5 & 64.8 & 58.9 \\
            \textbf{Exp2 E4M3} & 86.1 & 58.1 & 53.0 \\
            \textbf{Exp2 E5M2} & 87.8 & 62.4 & 58.5 \\
            \textbf{Exp2 E4M3xE5M2} & 87.5 & 63.5 & 57.4 \\
            \textbf{Exp2 HiF8} & 89.1 & 63.2 & 58.1 \\
            \textbf{E2E HiF8} & 88.0 & 64.0 & 58.3 \\
            \textbf{Naive E2E} & 87.8 & 54.0 & 44.0 \\
            \bottomrule
        \end{tabular}
    \end{center}
\end{table}

\begin{table}[htpb]
    \caption{Accuracy on different tasks in NLP With different setups. The accuracy is calculated in percentage and higher values are preferable.}
    \label{tab:scores_models}
    \vskip 0.15in
    \begin{center}
        \resizebox{\linewidth}{!}{
            \begin{tabular}{lccccc}
                \toprule
                \textbf{Model} & \textbf{Benchmark} & \textbf{Default} & \textbf{Exp2 HiF8} & \textbf{E2E HiF8} \\
                \midrule
                \multirow{3}{*}{\textbf{Qwen3}} & MMLU-P & 73.59 & 72.96 & 72.53 \\
                & GPQA-D & 54.55 & 57.58 & 58.59 \\
                & MATH500 & 78.40 & 80.00 & 79.80 \\
                \midrule
                \multirow{3}{*}{\textbf{Llama-3.1}} & MMLU-P & 46.01 & 45.51 & 45.19 \\
                & GPQA-D & 30.81 & 25.76 & 28.79 \\
                & MATH500 & 46.80 & 46.40 & 47.60 \\
                \bottomrule
            \end{tabular}
        }
    \end{center}
\end{table}

\subsection{The Multi-modal Models' Accuracy}

\begin{figure}[htpb]
    \renewcommand{\arraystretch}{0.5}
    \centering
    \begin{subfigure}{.45\linewidth}
        \centering
        \includegraphics[width=\linewidth]{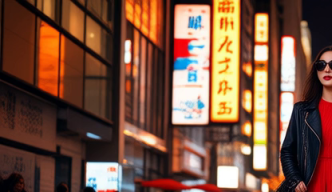}
        \caption{\textbf{Default}}
    \end{subfigure}
    \hfill
    \begin{subfigure}{.45\linewidth}
        \includegraphics[width=\linewidth]{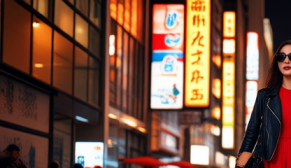}
        \caption{\textbf{Exp2 HiF8}}
    \end{subfigure}
    
    \begin{subfigure}{.45\linewidth}
        \centering
        \includegraphics[width=\linewidth]{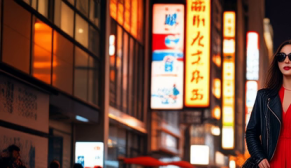}
        \caption{\textbf{E2E HiF8}}
    \end{subfigure}
    \hfill
    \begin{subfigure}{.45\linewidth}
        \includegraphics[width=\linewidth]{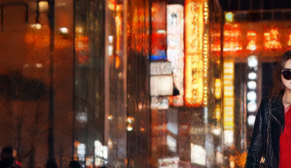}
        \caption{\textbf{Naive E2E}}
    \end{subfigure}
    \caption{The single frame generated from different setups using Wan-2.2. The illustration only shows the upper-left sub-frame for each setup. Only setup \textbf{Naive E2E} generated blurred frame and the other three frames are consistent with each other in a high quality.}
    \label{fig:wan_different_setups}
\end{figure}

We adopt the multi-modal model Wan-2.2~\cite{wan2025wan} as our testbed to evaluate the proposed method on the text-to-video (T2V) generation task.
Similar to the language models, the advantage of the proposed workflow can be easily verified with a simple experiment as shown in \Cref{fig:wan_different_setups}. This experiment only generates a single frame with forty sampling steps. Only setup \textbf{Naive E2E} generated a blurred frame, while the other three setups all generated consistent results in high quality.

8 prompts for T2V task is considered. The generated frames are shown in~\Cref{fig:prompt1}-\Cref{fig:prompt8}. To quantitatively analyze the quality of the generated videos, the average cosine similarity (Similarity), the average mean squared error (MSE), the average structural similarity index (SSIM), and the average peak signal-to-noise ratio (PSNR) across 8 generated videos are reported in \Cref{tab:wan-evaluation}. It compares the videos generated from setup \textbf{Exp2 HiF8} and \textbf{E2E HiF8} to \textbf{Default} setup, respectively. For both $\lambda = 1$ and $\lambda = 2$ shown in~\Cref{tab:wan-evaluation}, it shows only 1\% difference in the similarity between these two setups. This validates setup \textbf{E2E HiF8} is capable to be deployed for low-precision inference acceleration for harnessing extra savings on memory bandwidth consumption.

\begin{table}[htpb]
    \caption{Evaluation of Wan-2.2 on T2V tasks With different setups. Similarity, SSIM and PSNR are positively correlated with the generated contents' quality. MSE is negatively correlated with the generated contents' quality.}
    \label{tab:wan-evaluation}
    \vskip 0.15in
    \begin{center}
        %\begin{small}
            \begin{tabular}{lccc}
                \toprule
                \multirow{2}{*}{\textbf{Metric}} & \multirow{2}{*}{\textbf{Exp2 HiF8}} & \textbf{E2E HiF8} & \textbf{E2E HiF8} \\
                & & $\lambda = 1$ & $\lambda = 2$ \\
                \midrule
                Similarity($\uparrow$) & 0.9561 & 0.9503 & 0.9485 \\
                MSE($\downarrow$) & 1678.3 & 1826.13 & 1910.67 \\
                SSIM($\uparrow$) & 0.6511 & 0.6347 & 0.6225 \\
                PSNR($\uparrow$) & 19.446 & 18.875 & 18.389 \\
                \bottomrule
            \end{tabular}
        %\end{small}
    \end{center}
\end{table}

\subsection{The Restart Rate}

%Other than the quantitative metrics of model accuracy, we are also interested in the cost of restart in the tasks. This is because that any restart would invoke recalculation and passing of scores in high precision. This would break the pipeline of Flash-Attention and reduce the savings of time from low-precision exponentiation computations. To quantify the restart rate, we consider CWE-32K as the testbed for evaluating the restart rate in NLP tasks and using the first prompt shown in Appendix ~\Cref{fig:prompt1} in T2V task.
Other than the quantitative accuracy metrics, we also evaluate the cost associated with restart events. Each restart requires recalculation and transmission of scores in high precision, which breaks the Flash Attention pipeline and diminishes the time savings gained from low-precision exponentiation. To quantify the restart rate, we use the CWE-32K benchmark for NLP tasks and the first prompt from Appendix~\Cref{fig:prompt1} for the T2V task.

\begin{figure}[htbp]
    \centering
    \includegraphics[width=0.75\linewidth]{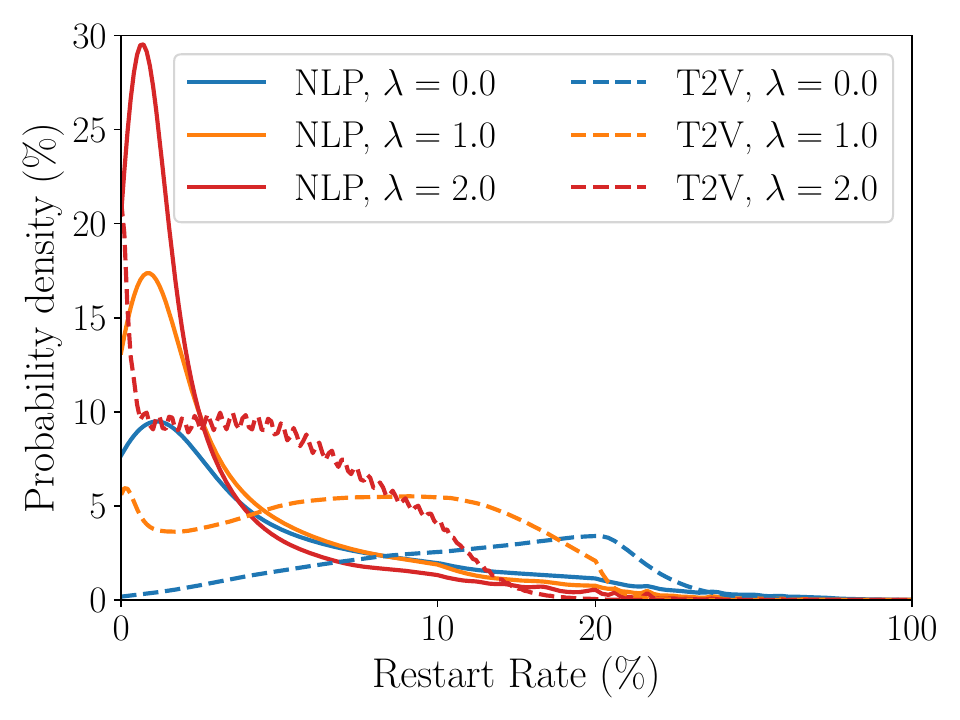}
    \caption{Probability density of restart rates for both NLP and T2V tasks. The solid line represents the density of NLP task and the dashed line represents the density of T2V task.}
    \label{fig:restart_analysis}
\end{figure}

%We count the number of restarted tiles in every run of \Cref{algo:workflow} and calculate the average restart rate (ARR) for the entire testbed problem. ARR represents the average probability of recalculation of $\mathbf{S}_i^{(j)}$ in high precision for each tile. We also record the peak restart rate (PRR) in any run of \Cref{algo:workflow} for the estimation of cost in the worst case. The overall restart rate is reported in \Cref{tab:restart-analysis} and the probability density of restart rates for every run of \Cref{algo:workflow} is shown in \Cref{fig:restart_analysis}. As it updates $\mathbf{d}_i^{(j)}$ and $\widetilde{\mathbf{O}}_i^{(j)}$ using integer powers of two for minimizing the accumulated error, the valid restart threshold $\lambda$ would only be integers. Therefore, we consider $\lambda = 0, 1, 2$ in the experiment.
We count the number of restarted tiles in each run of \Cref{algo:workflow} and compute the Average Restart Rate (ARR) for the entire testbed. ARR reflects the average probability that a tile's $\mathbf{S}_i^{(j)}$ requires high-precision recalculation. We also record the Peak Restart Rate (PRR) from any single calculation of $\mathbf{O}_i$ to estimate worst-case costs. The overall restart rates are presented in \Cref{tab:restart-analysis}, and the probability density of restart rates per run is shown in \Cref{fig:restart_analysis}. Since the method updates $\mathbf{d}_i^{(j)}$ and $\widetilde{\mathbf{O}}_i^{(j)}$ using integer powers of two to minimize accumulated error, the valid restart threshold $\lambda$ is constrained to integers. Thus, we experiment with $\lambda = 0, 1, 2$ in the experiment.

% Next, we focus on investigating the sensitivity of restart threshold $\lambda$ with respect to the generated content's quality. As shown in \Cref{tab:sensitivity-analysis}, choosing different restart thresholds $\lambda$ within ${0, 1,2}$ would not significantly influence the quality of the generated videos. 
%The different between setup \textbf{Exp2 HiF8} and \textbf{E2E HiF8} is only resulted from the re-ordering of transversal of tiles when calculating attentions. 
% By inspecting the generated frames as shown in {}, the major portion of the frames show a consistency while some of the minor details has been eliminated from the videos generated from setup \textbf{Exp2 HiF8} and \textbf{E2E HiF8}.

\begin{table}[thpb]
    \caption{Restart rate analysis of different restart thresholds $\lambda$. ARR denotes the average restart rate of each tile and PRR denotes the peak restart rate of any single calculation of $\mathbf{O}_i$.}
    \label{tab:restart-analysis}
    \vskip 0.15in
    \begin{center}
        \begin{small}
            \begin{tabular}{ccccc}
                \toprule
                \multirow{2}{*}{\textbf{Task Type}} & \multirow{2}{*}{\textbf{Metric}} & \multicolumn{3}{c}{$\lambda$} \\
                \cmidrule{3-5}
                & & 0.0 & 1.0 & 2.0 \\
                \midrule
                \multirow{2}{*}{NLP} & ARR & 10.86\% & 4.97\% & 2.50\% \\
                & PRR & 96.67\% & 91.30\% & 87.50\% \\
                \midrule
                \multirow{2}{*}{T2V} & ARR & 22.88\% & 10.48\% & 5.07\% \\
                % RR & 99.97\% & 94.95\% & 47.06\% \\
                & PRR & 87.89\% & 67.19\% & 61.72\% \\
                \bottomrule
            \end{tabular}
        \end{small}
    \end{center}
\end{table}

Compared with $\lambda = 0$, setting $\lambda = 1$ halves the ARR for both tasks. Given the negligible impact on model accuracy, \textbf{E2E HiF8} effectively harnesses the benefits of reduced memory bandwidth consumption, incurring only minor overhead from occasional high-precision recalculations. By tuning the restart threshold, it can also deliver a trade-off between models' accuracy and throughput. This tunability provides flexibility, enabling the proposed scheme to be adapted to systems with varying requirements. Besides ARR, PRR is another key indicator for evaluating the overhead of \textbf{E2E HiF8} setup. Although increasing the restart threshold does not sharply reduce the PRR as shown in \Cref{tab:restart-analysis}, \Cref{fig:restart_analysis} suggests that such high-restart cases are rare and thus unlikely to impact end-to-end inference throughput in practice. 
%In summary, this experiment solidifies the mitigation of the subtraction operation from the general-purpose accelerators to special-purpose accelerators for reducing the inference latency for different tasks.

\section{Conclusion}
\label{sec:conclusion}
In conclusion, this paper tackles the softmax bottleneck in Transformer attention by proposing a novel low-precision workflow based on 8-bit computation and block-aware rescaling. Our scheme addresses two key hardware limitations: (1) it halves the data movement bandwidth by constraining matrix multiplication outputs to 8-bit, and (2) it drastically reduces the area cost of exponentiation units by computing exponents in low (8-bit) precision. Evaluated across both language and multi-modal models, our method validates a concrete hardware-software co-design path, effectively doubling inference throughput potential without additional chip area. The approach is readily applicable to NLP and AIGC tasks with minimal adaptations. 
Future work will explore the interaction of our scheme with quantized attention inputs and its integration with lossless sparse attention methods like BLASST \cite{yuan2025blasst} for further gains. To maximize quality, integrating post-training techniques with our low-precision scheme is essential, as the model is agnostic to the arithmetic precision shift. Ultimately, extending this paradigm to the pre-training stage offers a promising avenue for reducing overall training costs.

%\section*{Impact Statement} This paper presents work whose goal is to advance the field of Machine Learning. There are many potential societal consequences of our work, none which we feel must be specifically highlighted here. 

\bibliography{ref}

@article{Ashish2017Attention,
  author       = {Ashish Vaswani and
                  Noam Shazeer and
                  Niki Parmar and
                  Jakob Uszkoreit and
                  Llion Jones and
                  Aidan N. Gomez and
                  Lukasz Kaiser and
                  Illia Polosukhin},
  title        = {Attention Is All You Need},
  journal      = {CoRR},
  volume       = {abs/1706.03762},
  year         = {2017},
  url          = {http://arxiv.org/abs/1706.03762},
  eprinttype    = {arXiv},
  eprint       = {1706.03762},
  bibsource    = {dblp computer science bibliography, https://dblp.org}
}

@misc{zhang2025sageattention,
      title={Sage{A}ttention: Accurate 8-Bit Attention for Plug-and-play Inference Acceleration}, 
      author={Jintao Zhang and Jia Wei and Haofeng Huang and Pengle Zhang and Jun Zhu and Jianfei Chen},
      year={2025},
      eprint={2410.02367},
      archivePrefix={arXiv},
      primaryClass={cs.LG},
      url={https://arxiv.org/abs/2410.02367}, 
}

@misc{nvidia2025pretraining,
      title={Pretraining Large Language Models with {NVFP4}}, 
      author={NVIDIA},
      year={2025},
      eprint={2509.25149},
      archivePrefix={arXiv},
      primaryClass={cs.CL},
      url={https://arxiv.org/abs/2509.25149}, 
}

@misc{wang2025sole,
      title={SOLE: Hardware-Software Co-design of Softmax and LayerNorm for Efficient Transformer Inference}, 
      author={Wenxun Wang and Shuchang Zhou and Wenyu Sun and Peiqin Sun and Yongpan Liu},
      year={2025},
      eprint={2510.17189},
      archivePrefix={arXiv},
      primaryClass={cs.LG},
      url={https://arxiv.org/abs/2510.17189}, 
}

@misc{alexandridis2025low,
      title={Low-Cost {F}lash{A}ttention with Fused Exponential and Multiplication Hardware Operators}, 
      author={Kosmas Alexandridis and Vasileios Titopoulos and Giorgos Dimitrakopoulos},
      year={2025},
      eprint={2505.14314},
      archivePrefix={arXiv},
      primaryClass={cs.AR},
      url={https://arxiv.org/abs/2505.14314}, 
}

@inproceedings{
    shah2024flashattention,
    title={Flash{A}ttention-3: Fast and Accurate Attention with Asynchrony and Low-precision},
    author={Jay Shah and Ganesh Bikshandi and Ying Zhang and Vijay Thakkar and Pradeep Ramani and Tri Dao},
    booktitle={The Thirty-eighth Annual Conference on Neural Information Processing Systems},
    year={2024},
    url={https://openreview.net/forum?id=tVConYid20}
}

@inproceedings{
    dao2024flashattention,
    title={Flash{A}ttention-2: Faster Attention with Better Parallelism and Work Partitioning},
    author={Tri Dao},
    booktitle={The Twelfth International Conference on Learning Representations},
    year={2024},
    url={https://openreview.net/forum?id=mZn2Xyh9Ec}
}

@misc{wan2025wan,
      title={Wan: Open and Advanced Large-Scale Video Generative Models}, 
      author={{Team Wan}},
      year={2025},
      eprint={2503.20314},
      archivePrefix={arXiv},
      primaryClass={cs.CV},
      url={https://arxiv.org/abs/2503.20314}, 
}

@article{Itay2018Quantized,
  author  = {Itay Hubara and Matthieu Courbariaux and Daniel Soudry and Ran El-Yaniv and Yoshua Bengio},
  title   = {Quantized Neural Networks: Training Neural Networks with Low Precision Weights and Activations},
  journal = {Journal of Machine Learning Research},
  year    = {2018},
  volume  = {18},
  number  = {187},
  pages   = {1--30},
  url     = {http://jmlr.org/papers/v18/16-456.html}
}

@misc{luo2024ascend,
      title={Ascend HiFloat8 Format for Deep Learning}, 
      author={Yuanyong Luo and Zhongxing Zhang and Richard Wu and Hu Liu and Ying Jin and Kai Zheng and Minmin Wang and Zhanying He and Guipeng Hu and Luyao Chen and Tianchi Hu and Junsong Wang and Minqi Chen and Mikhaylov Dmitry and Korviakov Vladimir and Bobrin Maxim and Yuhao Hu and Guanfu Chen and Zeyi Huang},
      year={2024},
      eprint={2409.16626},
      archivePrefix={arXiv},
      primaryClass={cs.LG},
      url={https://arxiv.org/abs/2409.16626}, 
}

@article{Frantar2022GPTQAP,
  title={GPTQ: Accurate Post-Training Quantization for Generative Pre-trained Transformers},
  author={Elias Frantar and Saleh Ashkboos and Torsten Hoefler and Dan Alistarh},
  journal={ArXiv},
  year={2022},
  volume={abs/2210.17323},
  url={https://api.semanticscholar.org/CorpusID:253237200}
}

@article{Lin2023AWQAW,
  title={AWQ: Activation-aware Weight Quantization for {LLM} Compression and Acceleration},
  author={Ji Lin and Jiaming Tang and Haotian Tang and Shang Yang and Xingyu Dang and Song Han},
  journal={ArXiv},
  year={2023},
  volume={abs/2306.00978},
  url={https://api.semanticscholar.org/CorpusID:271271084}
}

@INPROCEEDINGS{Stevens2021Softermax,
  author={Stevens, Jacob R. and Venkatesan, Rangharajan and Dai, Steve and Khailany, Brucek and Raghunathan, Anand},
  booktitle={2021 58th ACM/IEEE Design Automation Conference (DAC)}, 
  title={Softermax: Hardware/Software Co-Design of an Efficient Softmax for Transformers}, 
  year={2021},
  pages={469-474},
  doi={10.1109/DAC18074.2021.9586134}
}

@article{kaul2024attention,
  title={From attention to activation: Unravelling the enigmas of large language models},
  author={Kaul, Prannay and Ma, Chengcheng and Elezi, Ismail and Deng, Jiankang},
  journal={arXiv preprint arXiv:2410.17174},
  year={2024}
}

@book{bishop2006pattern,
  author    = {Bishop, Christopher M.},
  title     = {Pattern Recognition and Machine Learning},
  year      = {2006},
  publisher = {Springer},
  series    = {Information Science and Statistics},
  isbn      = {978-0387310732}
}

@misc{mishra2025recipe,
      title={Recipes for Pre-training {LLM}s with {MXFP8}}, 
      author={Asit Mishra and Dusan Stosic and Simon Layton and Paulius Micikevicius},
      year={2025},
      eprint={2506.08027},
      archivePrefix={arXiv},
      primaryClass={cs.LG},
      url={https://arxiv.org/abs/2506.08027}, 
}

@book{foundationsCVbook,
  title={Foundations of Computer Vision},
  author={Torralba, A. and Isola, P. and Freeman, W.T.},
  isbn={9780262378666},
  lccn={2023024589},
  series={Adaptive Computation and Machine Learning series},
  url={https://mitpress.mit.edu/9780262048972/foundations-of-computer-vision/},
  year={2024},
  publisher={MIT Press}
}

@book{Dongarra1998,
  author    = {Jack J. Dongarra and Iain S. Duff and Danny C. Sorensen and Henk A. van der Vorst},
  title     = {Numerical Linear Algebra for High-Performance Computers},
  publisher = {Society for Industrial and Applied Mathematics},
  year      = {1998},
  address   = {Philadelphia, PA},
  isbn      = {0-89871-428-1}
}

@book{murray1994mathematical,
  title={A Mathematical Introduction to Robotic Manipulation},
  author={Murray, Richard M. and Li, Zexiang and Sastry, S. Shankar},
  year={1994},
  publisher={CRC Press},
  isbn={9780849379819},
  address={Boca Raton, FL}
}

@misc{deepseekai2025deepseekv3,
      title={Deep{S}eek-V3 Technical Report}, 
      author={DeepSeek-AI},
      year={2025},
      eprint={2412.19437},
      archivePrefix={arXiv},
      primaryClass={cs.CL},
      url={https://arxiv.org/abs/2412.19437}, 
}

@misc{openai2024gpt4,
      title={{GPT}-4 Technical Report}, 
      author={OpenAI},
      year={2024},
      eprint={2303.08774},
      archivePrefix={arXiv},
      primaryClass={cs.CL},
      url={https://arxiv.org/abs/2303.08774}, 
}

@misc{hu2024case,
      title={Case-Based or Rule-Based: How Do Transformers Do the Math?}, 
      author={Yi Hu and Xiaojuan Tang and Haotong Yang and Muhan Zhang},
      year={2024},
      eprint={2402.17709},
      archivePrefix={arXiv},
      primaryClass={cs.AI},
      url={https://arxiv.org/abs/2402.17709}, 
}

@misc{sanghai2024advances,
      title={Advances in Transformers for Robotic Applications: A Review}, 
      author={Nikunj Sanghai and Nik Bear Brown},
      year={2024},
      eprint={2412.10599},
      archivePrefix={arXiv},
      primaryClass={cs.RO},
      url={https://arxiv.org/abs/2412.10599}, 
}

@misc{peebles2023scalable,
      title={Scalable Diffusion Models with Transformers}, 
      author={William Peebles and Saining Xie},
      year={2023},
      eprint={2212.09748},
      archivePrefix={arXiv},
      primaryClass={cs.CV},
      url={https://arxiv.org/abs/2212.09748}, 
}

@article{Wang2024Survey,
   title={A survey on large language model based autonomous agents},
   volume={18},
   ISSN={2095-2236},
   url={http://dx.doi.org/10.1007/s11704-024-40231-1},
   DOI={10.1007/s11704-024-40231-1},
   number={6},
   journal={Frontiers of Computer Science},
   publisher={Springer Science and Business Media LLC},
   author={Wang, Lei and Ma, Chen and Feng, Xueyang and Zhang, Zeyu and Yang, Hao and Zhang, Jingsen and Chen, Zhiyuan and Tang, Jiakai and Chen, Xu and Lin, Yankai and Zhao, Wayne Xin and Wei, Zhewei and Wen, Jirong},
   year={2024},
   month=mar }

@misc{qwen3technicalreport,
      title={Qwen3 Technical Report}, 
      author={Qwen Team},
      year={2025},
      eprint={2505.09388},
      archivePrefix={arXiv},
      primaryClass={cs.CL},
      url={https://arxiv.org/abs/2505.09388}, 
}

@misc{nvidiablackwellarichtecture,
    title={NVIDIA Blackwell Architecture Technical Brief},
    author={Nvidia},
    year={2025},
    url={https://resources.nvidia.com/en-us-blackwell-architecture}
}

@misc{yuan2025blasst,
      title={BLASST: Dynamic BLocked Attention Sparsity via Softmax Thresholding}, 
      author={Jiayi Yuan and Cameron Shinn and Kai Xu and Jingze Cui and George Klimiashvili and Guangxuan Xiao and Perkz Zheng and Bo Li and Yuxin Zhou and Zhouhai Ye and Weijie You and Tian Zheng and Dominic Brown and Pengbo Wang and Richard Cai and Julien Demouth and John D. Owens and Xia Hu and Song Han and Timmy Liu and Huizi Mao},
      year={2025},
      eprint={2512.12087},
      archivePrefix={arXiv},
      primaryClass={cs.CL},
      url={https://arxiv.org/abs/2512.12087}, 
}

@misc{wang2024mmlupro,
      title={{MMLU}-{P}ro: A More Robust and Challenging Multi-Task Language Understanding Benchmark}, 
      author={Yubo Wang and Xueguang Ma and Ge Zhang and Yuansheng Ni and Abhranil Chandra and Shiguang Guo and Weiming Ren and Aaran Arulraj and Xuan He and Ziyan Jiang and Tianle Li and Max Ku and Kai Wang and Alex Zhuang and Rongqi Fan and Xiang Yue and Wenhu Chen},
      year={2024},
      eprint={2406.01574},
      archivePrefix={arXiv},
      primaryClass={cs.CL},
      url={https://arxiv.org/abs/2406.01574}, 
}

@INPROCEEDINGS{Liao2019DaVinci,
  author={Liao, Heng and Tu, Jiajin and Xia, Jing and Zhou, Xiping},
  booktitle={2019 IEEE Hot Chips 31 Symposium (HCS)}, 
  title={DaVinci: A Scalable Architecture for Neural Network Computing}, 
  year={2019},
  volume={},
  number={},
  pages={1-44},
  doi={10.1109/HOTCHIPS.2019.8875654}
}

@misc{grattafiori2024llama3,
      title={The {L}lama 3 Herd of Models}, 
      author={Llama Team},
      year={2024},
      eprint={2407.21783},
      archivePrefix={arXiv},
      primaryClass={cs.AI},
      url={https://arxiv.org/abs/2407.21783}, 
}

@misc{lightman2023lets,
      title={Let's Verify Step by Step}, 
      author={Hunter Lightman and Vineet Kosaraju and Yura Burda and Harri Edwards and Bowen Baker and Teddy Lee and Jan Leike and John Schulman and Ilya Sutskever and Karl Cobbe},
      year={2023},
      eprint={2305.20050},
      archivePrefix={arXiv},
      primaryClass={cs.LG},
      url={https://arxiv.org/abs/2305.20050}, 
}

@article{hsieh2024ruler,
  title={{RULER}: What's the Real Context Size of Your Long-Context Language Models?},
  author={Cheng-Ping Hsieh and Simeng Sun and Samuel Kriman and Shantanu Acharya and Dima Rekesh and Fei Jia and Yang Zhang and Boris Ginsburg},
  year={2024},
  journal={arXiv preprint arXiv:2404.06654},
}

@misc{rein2023gpqa,
      title={{GPQA}: A Graduate-Level {G}oogle-{P}roof {Q\&A} Benchmark}, 
      author={David Rein and Betty Li Hou and Asa Cooper Stickland and Jackson Petty and Richard Yuanzhe Pang and Julien Dirani and Julian Michael and Samuel R. Bowman},
      year={2023},
      eprint={2311.12022},
      archivePrefix={arXiv},
      primaryClass={cs.AI},
      url={https://arxiv.org/abs/2311.12022}, 
}
\bibliographystyle{icml2026}

%%%%%%%%%%%%%%%%%%%%%%%%%%%%%%%%%%%%%%%%%%%%%%%%%%%%%%%%%%%%%%%%%%%%%%%%%%%%%%%
%%%%%%%%%%%%%%%%%%%%%%%%%%%%%%%%%%%%%%%%%%%%%%%%%%%%%%%%%%%%%%%%%%%%%%%%%%%%%%%
% APPENDIX
%%%%%%%%%%%%%%%%%%%%%%%%%%%%%%%%%%%%%%%%%%%%%%%%%%%%%%%%%%%%%%%%%%%%%%%%%%%%%%%
%%%%%%%%%%%%%%%%%%%%%%%%%%%%%%%%%%%%%%%%%%%%%%%%%%%%%%%%%%%%%%%%%%%%%%%%%%%%%%%
\newpage
\appendix
\onecolumn

\section{Implementation of 8-bit Operations}

\subsection{HiF8 Implementation}

To simplify the implementation of the experiments, this paper consistently uses HiF8 \cite{luo2024ascend} as the unified 8-bit number format. As indicated in \cite{luo2024ascend}, HiF8 can harness the advantage of E4M3 and E5M2 with a wider dynamic range for a best-of-both-worlds representation. With no prior knowledge of the dynamic ranges of scores in the model and their exponents, HiF8 is expected to be much more robust compared to other FP8 formats. As no hardware implementation of powers of two in 8-bit numbers is currently available, the utilization of HiF8 operations is implemented in a simulated manner by taking advantage of FP16 operations. All of the inputs are truncated to HiF8 and then retained back to FP16 for following steps, as shown in line 2 and 3 in \Cref{algo:exp2_hif8}. Then the required operations, e.g., powers of 2, are performed in FP16. Finally, the numbers are converted to HiF8 to truncate the extra bits provided by FP16. This procedure is summarized in \Cref{algo:exp2_hif8}.

\begin{algorithm}[htb]
    \caption{Simulation of Powers of Two in HiF8.}
    \label{algo:exp2_hif8}
    \begin{algorithmic}[1]
        \STATE {\bfseries Input:} data $x_{\mathrm{FP16}}$ in FP16.
        \STATE Convert $x_{\mathrm{FP16}}$ to HiF8 $x_{\mathrm{HiF8}}$.
        \STATE Convert $x_{\mathrm{HiF8}}$ to FP16 $\hat{x}_{\mathrm{FP16}}$.
        \STATE Compute powers of two $y_{\mathrm{FP16}} = \mathrm{exp2}(\hat{x}_{\mathrm{HiF8}})$
        \STATE Convert $y_{\mathrm{FP16}}$ to HiF8 $y_{\mathrm{HiF8}}$.
        \STATE Convert $y_{\mathrm{HiF8}}$ to FP16 $\hat{y}_{\mathrm{FP16}}$.
        \STATE {\bfseries Return:} $\hat{y}_{\mathrm{FP16}}$.
    \end{algorithmic}
\end{algorithm}

\section{Ablation Experiments Results}

\subsection{The Language Models}

\Cref{tab:mmlu-qwen3} and \ref{tab:mmlu-llama} show the scores of Qwen3 and Llama-3.1 scores on different sub-tasks in MMLU-Pro benchmark suite, respectively. All of the accuracies that showing a degradation more than 1\% (at least two-sample differences) of the task have been highlighted in red.

\begin{table}[htp]
    \caption{Accuracy of Qwen3-30B-A3B on different tasks in MMLU-Pro with different setups. The accuracy is calculated in percentage and higher values are preferable.}
    \label{tab:mmlu-qwen3}
    \vskip 0.15in
    \begin{center}
        \begin{tabular}{lccc}
            \toprule
            \diagbox{Task}{Setup} & \textbf{Default} & \textbf{Exp2 HiF8} & \textbf{E2E HiF8} \\
            \midrule
            Computer Science & 78.29 & \textcolor{red}{76.34} & 77.32 \\
            Math & 85.79 & 85.34 & 84.97 \\
            Chemistry & 76.86 & 76.94 & 76.24 \\
            Engineering & 54.39 & \textcolor{red}{51.70} & \textcolor{red}{51.60} \\
            Law & 49.59 & 49.23 & 49.59 \\
            Biology & 87.31 & 86.33 & 86.75 \\
            Health & 74.57 & 74.69 & 74.08 \\
            Physics & 78.37 & 78.44 & 77.75 \\
            Business & 76.93 & 77.69 & 76.17 \\
            Philosophy & 69.94 & \textcolor{red}{66.93} & \textcolor{red}{68.34} \\
            Economics & 82.35 & \textcolor{red}{81.28} & 81.64 \\
            Other & 70.35 & 70.78 & 70.35 \\
            Psychology & 78.20 & 77.69 & 77.32 \\
            History & 66.14 & \textcolor{red}{63.78} & \textcolor{red}{63.25} \\
            \midrule
            Overall & 73.59 & 72.96 & 72.53 \\
            \bottomrule
        \end{tabular}
    \end{center}
\end{table}

\begin{table}[htp]
    \caption{Accuracy of Llama-3.1-8B on different tasks in MMLU-Pro with different setups. The accuracy is calculated in percentage and higher values are preferable.}
    \label{tab:mmlu-llama}
    \vskip 0.15in
    \begin{center}
        \begin{tabular}{lccc}
            \toprule
            \diagbox{Task}{Setup} & \textbf{Default} & \textbf{Exp2 HiF8} & \textbf{E2E HiF8} \\
            \midrule
            Computer Science & 48.54 & \textcolor{red}{45.85} & 48.78 \\
            Math & 47.30 & 49.15 & 48.85 \\
            Chemistry & 40.46 & \textcolor{red}{37.72} & \textcolor{red}{37.46} \\
            Engineering & 25.39 & \textcolor{red}{23.84} & \textcolor{red}{23.32} \\
            Law & 32.43 & \textcolor{red}{31.06} & \textcolor{red}{31.34} \\
            Biology & 64.16 & 66.67 & 66.11 \\
            Health & 56.11 & 55.62 & 56.60 \\
            Physics & 39.65 & 39.26 & 39.18 \\
            Business & 51.71 & 50.95 & \textcolor{red}{49.68} \\
            Philosophy & 54.27 & \textcolor{red}{43.29} & \textcolor{red}{42.69} \\
            Economics & 49.68 & 55.57 & 53.67 \\
            Other & 49.68 & 49.35 & 46.75 \\
            Psychology & 61.15 & \textcolor{red}{60.03} & 61.15 \\
            History & 43.04 & \textcolor{red}{41.73} & \textcolor{red}{41.47} \\
            \midrule
            Overall & 46.01 & 45.51 & 45.19 \\
            \bottomrule
        \end{tabular}
    \end{center}
\end{table}

\Cref{tab:math500-qwen3} and \ref{tab:math500-llama} show the scores of Qwen3 and Llama-3.1 scores on different difficulty levels in MATH500.

\begin{table}[htp]
    \caption{Accuracy of Qwen3-30B-A3B on different difficulty levels in MATH500 with different setups. The accuracy is calculated in percentage and higher values are preferable.}
    \label{tab:math500-qwen3}
    \vskip 0.15in
    \begin{center}
        \begin{tabular}{cccc}
            \toprule
            \diagbox{Level}{Setup} & \textbf{Default} & \textbf{Exp2 HiF8} & \textbf{E2E HiF8} \\
            \midrule
            1 & 97.67 & 97.67 & 95.35 \\
            2 & 92.22 & 93.33 & 96.67 \\
            3 & 86.67 & 90.48 & 87.62 \\
            4 & 75.78 & 77.34 & 75.78 \\
            5 & 58.96 & 59.70 & 61.19 \\
            \midrule
            Overall & 78.4 & 80.0 & 79.8 \\
            \bottomrule
        \end{tabular}
    \end{center}
\end{table}

\begin{table}[htp]
    \caption{Scores of Llama-3.1-8B on different difficulty levels in MATH500 with different setups. The accuracy is calculated in percentage and higher values are preferable.}
    \label{tab:math500-llama}
    \vskip 0.15in
    \begin{center}
        \begin{tabular}{cccc}
            \toprule
            \diagbox{Level}{Setup} & \textbf{Default} & \textbf{Exp2 HiF8} & \textbf{E2E HiF8} \\
            \midrule
            1 & 83.72 & 86.05 & 93.02 \\
            2 & 68.89 & 71.11 & 64.44 \\
            3 & 53.33 & 58.10 & 57.14 \\
            4 & 42.19 & 34.38 & 42.19 \\
            5 & 19.40 & 19.40 & 19.40 \\
            \midrule
            Overall & 46.8 & 46.4 & 47.6 \\
            \bottomrule
        \end{tabular}
    \end{center}
\end{table}

\subsection{The Multi-modal Models}

\begin{figure}[htpb]
    \renewcommand{\arraystretch}{0.5}
    \setlength{\tabcolsep}{0pt}
    \centering
    \begin{tabular}{ccccc}
        \centering
        \adjustbox{valign=c, rotate=90}{\centering \textbf{\quad\  Default}} & \includegraphics[width=.2\linewidth]{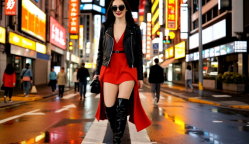} & \includegraphics[width=.2\linewidth]{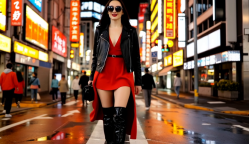} & \includegraphics[width=.2\linewidth]{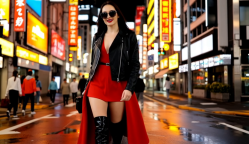} & \includegraphics[width=.2\linewidth]{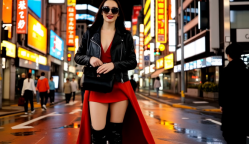} \\
        \adjustbox{valign=c, rotate=90}{\centering \textbf{\ \ \  Exp2 HiF8}} & \includegraphics[width=.2\linewidth]{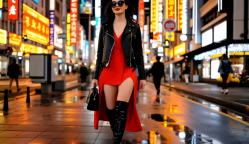} & \includegraphics[width=.2\linewidth]{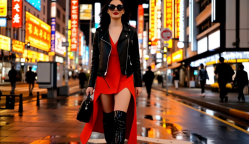} & \includegraphics[width=.2\linewidth]{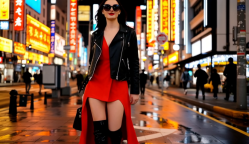} & \includegraphics[width=.2\linewidth]{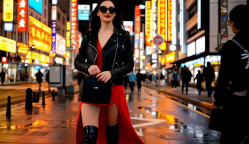} \\
        \adjustbox{valign=c, rotate=90}{\begin{minipage}{2cm} \centering \textbf{E2E HiF8} \\ $\lambda = 1$\end{minipage}} & \includegraphics[width=.2\linewidth]{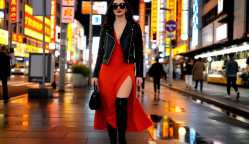} & \includegraphics[width=.2\linewidth]{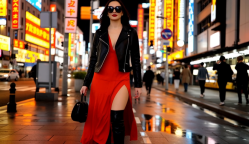} & \includegraphics[width=.2\linewidth]{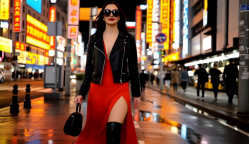} & \includegraphics[width=.2\linewidth]{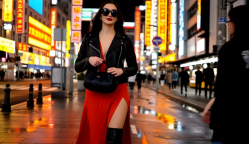} \\
        \adjustbox{valign=c, rotate=90}{\begin{minipage}{2cm} \centering \textbf{E2E HiF8} \\ $\lambda = 2$\end{minipage}} & \includegraphics[width=.2\linewidth]{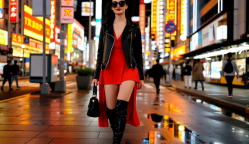} & \includegraphics[width=.2\linewidth]{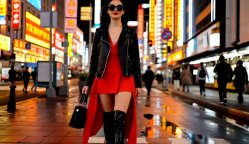} & \includegraphics[width=.2\linewidth]{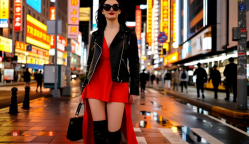} & \includegraphics[width=.2\linewidth]{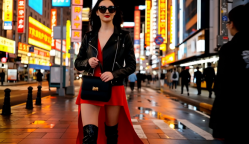} \\
    \end{tabular}
    \caption{Prompt 1: ``\textit{A stylish woman walks down a Tokyo street filled with warm glowing neon and animated city signage. She wears a black leather jacket, a long red dress, and black boots, and carries a black purse. She wears sunglasses and red lipstick. She walks confidently and casually. The street is damp and reflective, creating a mirror effect of the colorful lights. Many pedestrians walk about.}''}
    \label{fig:prompt1}
\end{figure}

\begin{figure}[htpb]
    \renewcommand{\arraystretch}{0.5}
    \setlength{\tabcolsep}{0pt}
    \centering
    \begin{tabular}{ccccc}
        \centering
        \adjustbox{valign=c, rotate=90}{\centering \textbf{\quad\  Default}} & \includegraphics[width=.2\linewidth]{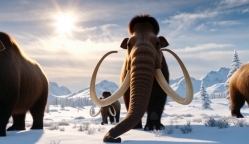} & \includegraphics[width=.2\linewidth]{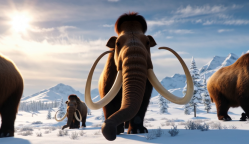} & \includegraphics[width=.2\linewidth]{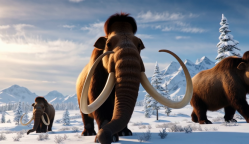} & \includegraphics[width=.2\linewidth]{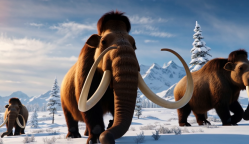} \\
        \adjustbox{valign=c, rotate=90}{\centering \textbf{\ \ \  Exp2 HiF8}} & \includegraphics[width=.2\linewidth]{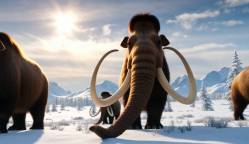} & \includegraphics[width=.2\linewidth]{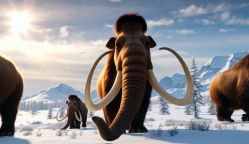} & \includegraphics[width=.2\linewidth]{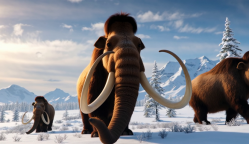} & \includegraphics[width=.2\linewidth]{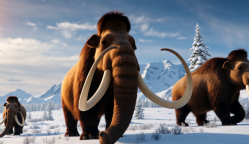} \\
        \adjustbox{valign=c, rotate=90}{\begin{minipage}{2cm} \centering \textbf{E2E HiF8} \\ $\lambda = 1$\end{minipage}} & \includegraphics[width=.2\linewidth]{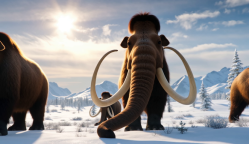} & \includegraphics[width=.2\linewidth]{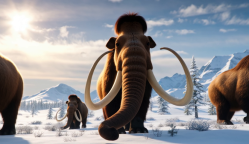} & \includegraphics[width=.2\linewidth]{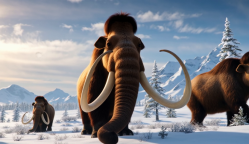} & \includegraphics[width=.2\linewidth]{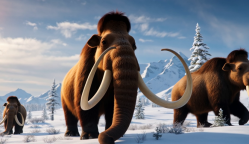} \\
        \adjustbox{valign=c, rotate=90}{\begin{minipage}{2cm} \centering \textbf{E2E HiF8} \\ $\lambda = 2$\end{minipage}} & \includegraphics[width=.2\linewidth]{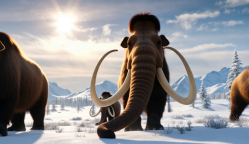} & \includegraphics[width=.2\linewidth]{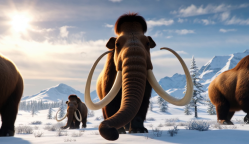} & \includegraphics[width=.2\linewidth]{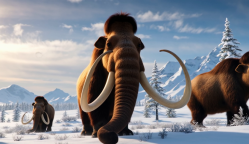} & \includegraphics[width=.2\linewidth]{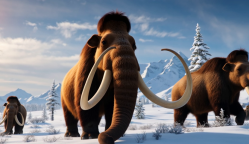} \\
    \end{tabular}
    \caption{Prompt 2: ``\textit{Several giant wooly mammoths approach treading through a snowy meadow, their long wooly fur lightly blows in the wind as they walk, snow covered trees and dramatic snow capped mountains in the distance, mid afternoon light with wispy clouds and a sun high in the distance creates a warm glow, the low camera view is stunning capturing the large furry mammal with beautiful photography, depth of field.}''}
    \label{fig:prompt2}
\end{figure}

\begin{figure}[htp]
    \renewcommand{\arraystretch}{0.5}
    \setlength{\tabcolsep}{0pt}
    \centering
    \begin{tabular}{ccccc}
        \centering
        \adjustbox{valign=c, rotate=90}{\centering \textbf{\quad\  Default}} & \includegraphics[width=.2\linewidth]{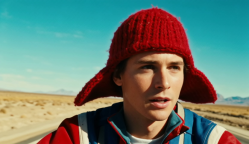} & \includegraphics[width=.2\linewidth]{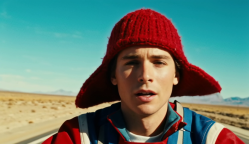} & \includegraphics[width=.2\linewidth]{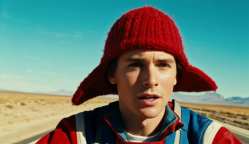} & \includegraphics[width=.2\linewidth]{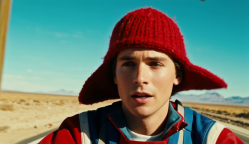} \\
        \adjustbox{valign=c, rotate=90}{\centering \textbf{\ \ \  Exp2 HiF8}} & \includegraphics[width=.2\linewidth]{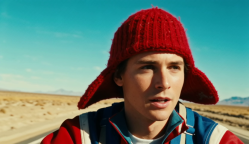} & \includegraphics[width=.2\linewidth]{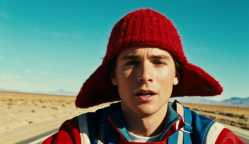} & \includegraphics[width=.2\linewidth]{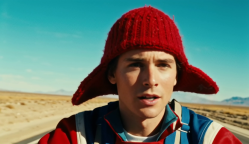} & \includegraphics[width=.2\linewidth]{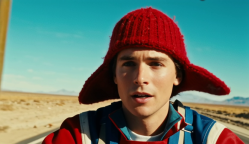} \\
        \adjustbox{valign=c, rotate=90}{\begin{minipage}{2cm} \centering \textbf{E2E HiF8} \\ $\lambda = 1$\end{minipage}} & \includegraphics[width=.2\linewidth]{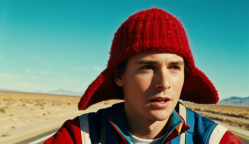} & \includegraphics[width=.2\linewidth]{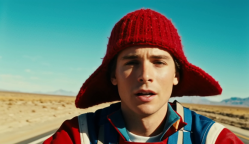} & \includegraphics[width=.2\linewidth]{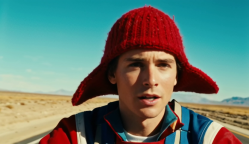} & \includegraphics[width=.2\linewidth]{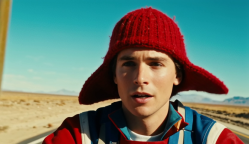} \\
        \adjustbox{valign=c, rotate=90}{\begin{minipage}{2cm} \centering \textbf{E2E HiF8} \\ $\lambda = 2$\end{minipage}} & \includegraphics[width=.2\linewidth]{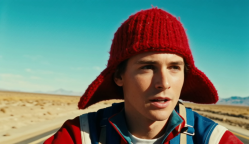} & \includegraphics[width=.2\linewidth]{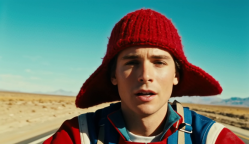} & \includegraphics[width=.2\linewidth]{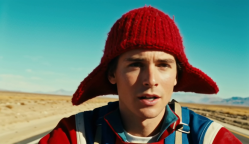} & \includegraphics[width=.2\linewidth]{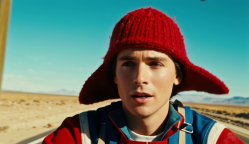} \\
    \end{tabular}
    \caption{Prompt 3: ``\textit{A movie trailer featuring the adventures of the 30 year old space man wearing a red wool knitted motorcycle helmet, blue sky, salt desert, cinematic style, shot on 35mm film, vivid colors.}''}
    \label{fig:prompt3}
\end{figure}

\begin{figure}[htp]
    \renewcommand{\arraystretch}{0.5}
    \setlength{\tabcolsep}{0pt}
    \centering
    \begin{tabular}{ccccc}
        \centering
        \adjustbox{valign=c, rotate=90}{\centering \textbf{\quad\  Default}} & \includegraphics[width=.2\linewidth]{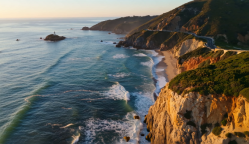} & \includegraphics[width=.2\linewidth]{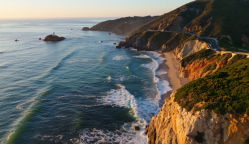} & \includegraphics[width=.2\linewidth]{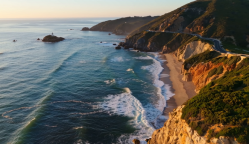} & \includegraphics[width=.2\linewidth]{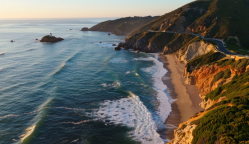} \\
        \adjustbox{valign=c, rotate=90}{\centering \textbf{\ \ \  Exp2 HiF8}} & \includegraphics[width=.2\linewidth]{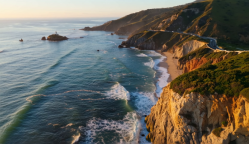} & \includegraphics[width=.2\linewidth]{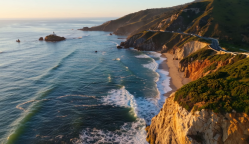} & \includegraphics[width=.2\linewidth]{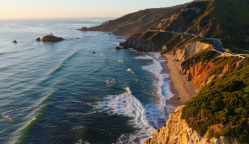} & \includegraphics[width=.2\linewidth]{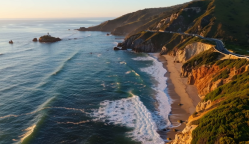} \\
        \adjustbox{valign=c, rotate=90}{\begin{minipage}{2cm} \centering \textbf{E2E HiF8} \\ $\lambda = 1$\end{minipage}} & \includegraphics[width=.2\linewidth]{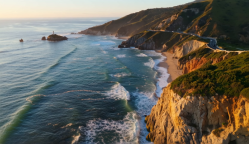} & \includegraphics[width=.2\linewidth]{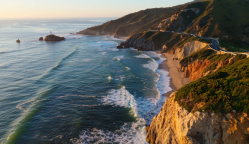} & \includegraphics[width=.2\linewidth]{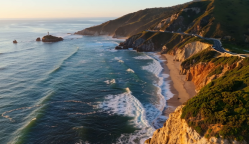} & \includegraphics[width=.2\linewidth]{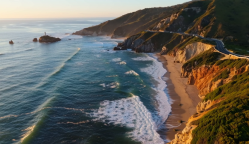} \\
        \adjustbox{valign=c, rotate=90}{\begin{minipage}{2cm} \centering \textbf{E2E HiF8} \\ $\lambda = 2$\end{minipage}} & \includegraphics[width=.2\linewidth]{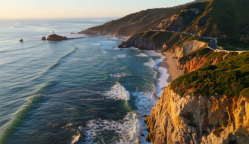} & \includegraphics[width=.2\linewidth]{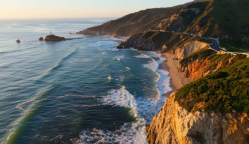} & \includegraphics[width=.2\linewidth]{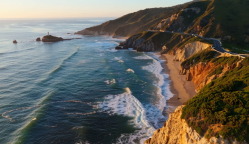} & \includegraphics[width=.2\linewidth]{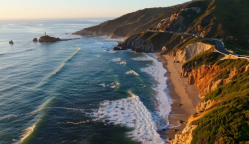} \\
    \end{tabular}
    \caption{Prompt 4  ``\textit{Drone view of waves crashing against the rugged cliffs along Big Sur's garay point beach. The crashing blue waters create white-tipped waves, while the golden light of the setting sun illuminates the rocky shore. A small island with a lighthouse sits in the distance, and green shrubbery covers the cliff's edge. The steep drop from the road down to the beach is a dramatic feat, with the cliff's edges jutting out over the sea. This is a view that captures the raw beauty of the coast and the rugged landscape of the Pacific Coast Highway.}''}
    \label{fig:prompt4}
\end{figure}

\begin{figure}[htp]
    \renewcommand{\arraystretch}{0.5}
    \setlength{\tabcolsep}{0pt}
    \centering
    \begin{tabular}{ccccc}
        \centering
        \adjustbox{valign=c, rotate=90}{\centering \textbf{\quad\  Default}} & \includegraphics[width=.2\linewidth]{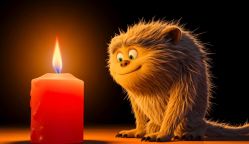} & \includegraphics[width=.2\linewidth]{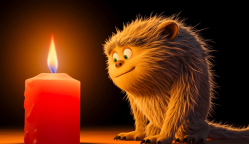} & \includegraphics[width=.2\linewidth]{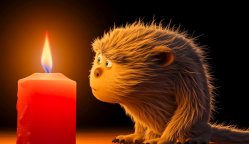} & \includegraphics[width=.2\linewidth]{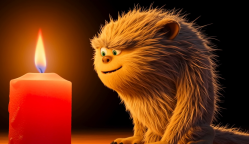} \\
        \adjustbox{valign=c, rotate=90}{\centering \textbf{\ \ \  Exp2 HiF8}} & \includegraphics[width=.2\linewidth]{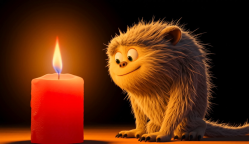} & \includegraphics[width=.2\linewidth]{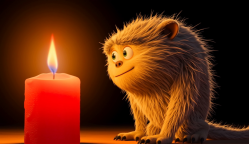} & \includegraphics[width=.2\linewidth]{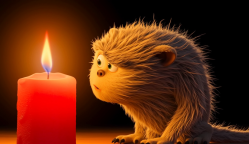} & \includegraphics[width=.2\linewidth]{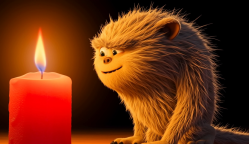} \\
        \adjustbox{valign=c, rotate=90}{\begin{minipage}{2cm} \centering \textbf{E2E HiF8} \\ $\lambda = 1$\end{minipage}} & \includegraphics[width=.2\linewidth]{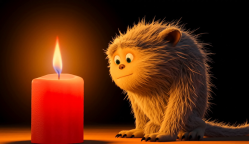} & \includegraphics[width=.2\linewidth]{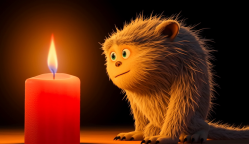} & \includegraphics[width=.2\linewidth]{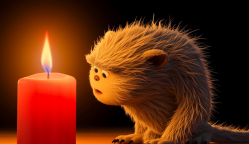} & \includegraphics[width=.2\linewidth]{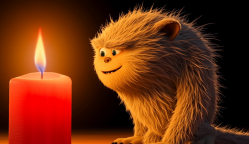} \\
        \adjustbox{valign=c, rotate=90}{\begin{minipage}{2cm} \centering \textbf{E2E HiF8} \\ $\lambda = 2$\end{minipage}} & \includegraphics[width=.2\linewidth]{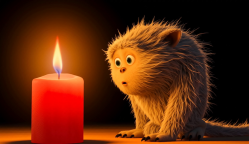} & \includegraphics[width=.2\linewidth]{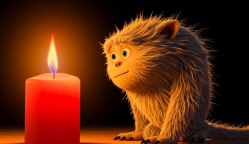} & \includegraphics[width=.2\linewidth]{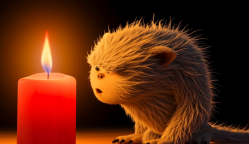} & \includegraphics[width=.2\linewidth]{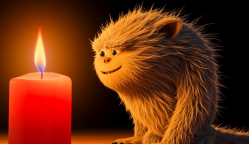} \\
    \end{tabular}
    \caption{Prompt 5: ``\textit{Animated scene features a close-up of a short fluffy monster kneeling beside a melting red candle. The art style is 3D and realistic, with a focus on lighting and texture. The mood of the painting is one of wonder and curiosity, as the monster gazes at the flame with wide eyes and open mouth. Its pose and expression convey a sense of innocence and playfulness, as if it is exploring the world around it for the first time. The use of warm colors and dramatic lighting further enhances the cozy atmosphere of the image.}''}
    \label{fig:prompt5}
\end{figure}

\begin{figure}[htp]
    \renewcommand{\arraystretch}{0.5}
    \setlength{\tabcolsep}{0pt}
    \centering
    \begin{tabular}{ccccc}
        \centering
        \adjustbox{valign=c, rotate=90}{\centering \textbf{\quad\  Default}} & \includegraphics[width=.2\linewidth]{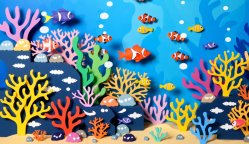} & \includegraphics[width=.2\linewidth]{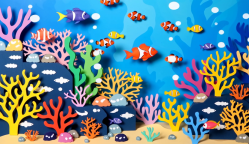} & \includegraphics[width=.2\linewidth]{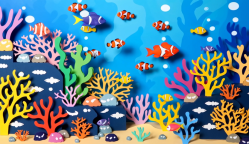} & \includegraphics[width=.2\linewidth]{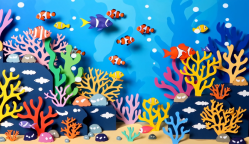} \\
        \adjustbox{valign=c, rotate=90}{\centering \textbf{\ \ \  Exp2 HiF8}} & \includegraphics[width=.2\linewidth]{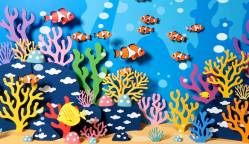} & \includegraphics[width=.2\linewidth]{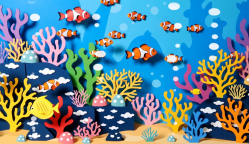} & \includegraphics[width=.2\linewidth]{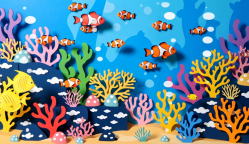} & \includegraphics[width=.2\linewidth]{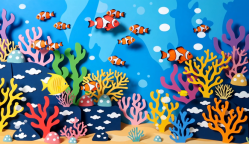} \\
        \adjustbox{valign=c, rotate=90}{\begin{minipage}{2cm} \centering \textbf{E2E HiF8} \\ $\lambda = 1$\end{minipage}} & \includegraphics[width=.2\linewidth]{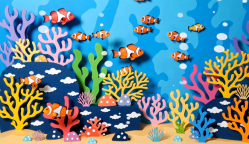} & \includegraphics[width=.2\linewidth]{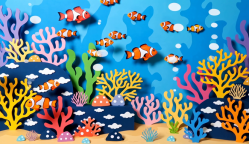} & \includegraphics[width=.2\linewidth]{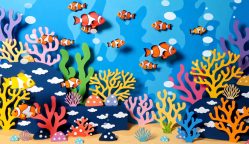} & \includegraphics[width=.2\linewidth]{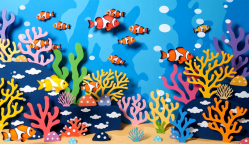} \\
        \adjustbox{valign=c, rotate=90}{\begin{minipage}{2cm} \centering \textbf{E2E HiF8} \\ $\lambda = 2$\end{minipage}} & \includegraphics[width=.2\linewidth]{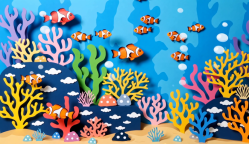} & \includegraphics[width=.2\linewidth]{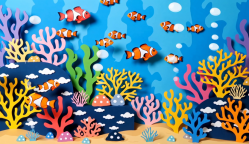} & \includegraphics[width=.2\linewidth]{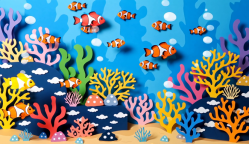} & \includegraphics[width=.2\linewidth]{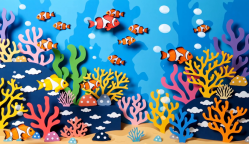} \\
    \end{tabular}
    \caption{Prompt 6: ``\textit{A gorgeously rendered papercraft world of a coral reef, rife with colorful fish and sea creatures.}''}
    \label{fig:prompt6}
\end{figure}

\begin{figure}
    \renewcommand{\arraystretch}{0.5}
    \setlength{\tabcolsep}{0pt}
    \centering
    \begin{tabular}{ccccc}
        \centering
        \adjustbox{valign=c, rotate=90}{\centering \textbf{\quad\  Default}} & \includegraphics[width=.2\linewidth]{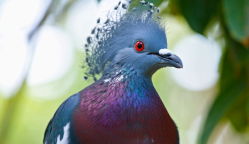} & \includegraphics[width=.2\linewidth]{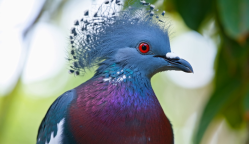} & \includegraphics[width=.2\linewidth]{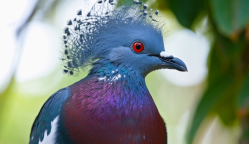} & \includegraphics[width=.2\linewidth]{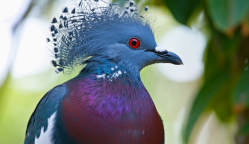} \\
        \adjustbox{valign=c, rotate=90}{\centering \textbf{\ \ \  Exp2 HiF8}} & \includegraphics[width=.2\linewidth]{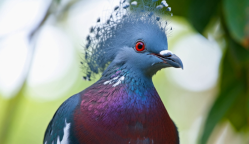} & \includegraphics[width=.2\linewidth]{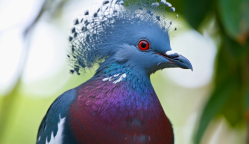} & \includegraphics[width=.2\linewidth]{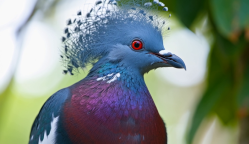} & \includegraphics[width=.2\linewidth]{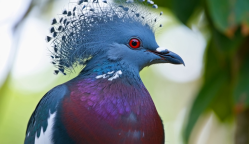} \\
        \adjustbox{valign=c, rotate=90}{\begin{minipage}{2cm} \centering \textbf{E2E HiF8} \\ $\lambda = 1$\end{minipage}} & \includegraphics[width=.2\linewidth]{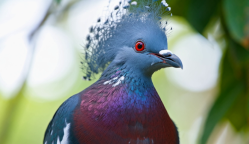} & \includegraphics[width=.2\linewidth]{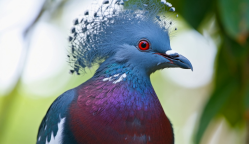} & \includegraphics[width=.2\linewidth]{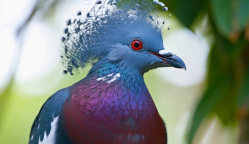} & \includegraphics[width=.2\linewidth]{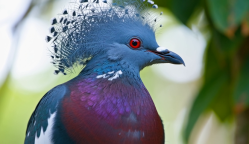} \\
        \adjustbox{valign=c, rotate=90}{\begin{minipage}{2cm} \centering \textbf{E2E HiF8} \\ $\lambda = 2$\end{minipage}} & \includegraphics[width=.2\linewidth]{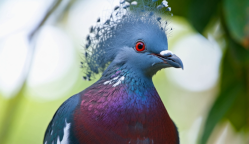} & \includegraphics[width=.2\linewidth]{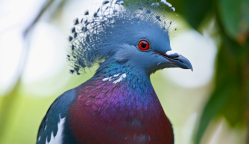} & \includegraphics[width=.2\linewidth]{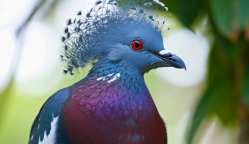} & \includegraphics[width=.2\linewidth]{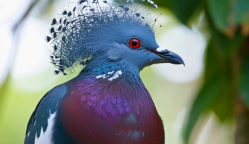} \\
    \end{tabular}
    \caption{Prompt 7: ``\textit{This close-up shot of a Victoria crowned pigeon showcases its striking blue plumage and red chest. Its crest is made of delicate, lacy feathers, while its eye is a striking red color. The bird's head is tilted slightly to the side, giving the impression of it looking regal and majestic. The background is blurred, drawing attention to the bird's striking appearance.}''}
    \label{fig:prompt7}
\end{figure}

\begin{figure}[htp]
    \renewcommand{\arraystretch}{0.5}
    \setlength{\tabcolsep}{0pt}
    \centering
    \begin{tabular}{ccccc}
        \centering
        \adjustbox{valign=c, rotate=90}{\centering \textbf{\quad\  Default}} & \includegraphics[width=.2\linewidth]{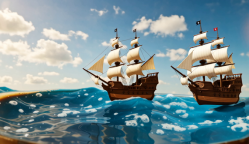} & \includegraphics[width=.2\linewidth]{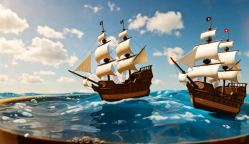} & \includegraphics[width=.2\linewidth]{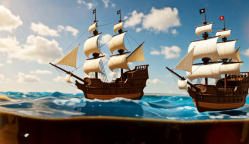} & \includegraphics[width=.2\linewidth]{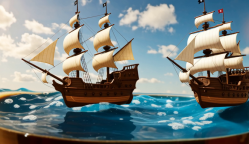} \\
        \adjustbox{valign=c, rotate=90}{\centering \textbf{\ \ \  Exp2 HiF8}} & \includegraphics[width=.2\linewidth]{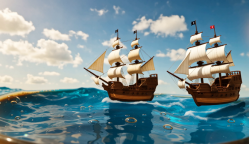} & \includegraphics[width=.2\linewidth]{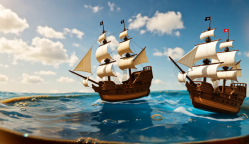} & \includegraphics[width=.2\linewidth]{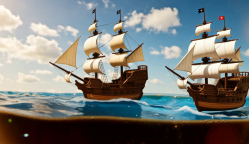} & \includegraphics[width=.2\linewidth]{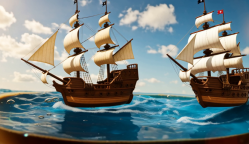} \\
        \adjustbox{valign=c, rotate=90}{\begin{minipage}{2cm} \centering \textbf{E2E HiF8} \\ $\lambda = 1$\end{minipage}} & \includegraphics[width=.2\linewidth]{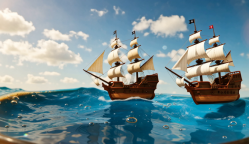} & \includegraphics[width=.2\linewidth]{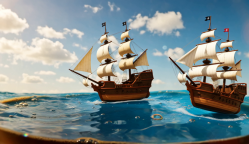} & \includegraphics[width=.2\linewidth]{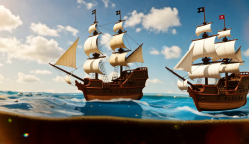} & \includegraphics[width=.2\linewidth]{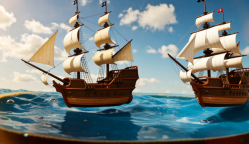} \\
        \adjustbox{valign=c, rotate=90}{\begin{minipage}{2cm} \centering \textbf{E2E HiF8} \\ $\lambda = 2$\end{minipage}} & \includegraphics[width=.2\linewidth]{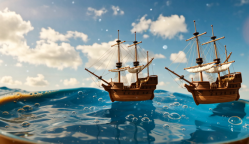} & \includegraphics[width=.2\linewidth]{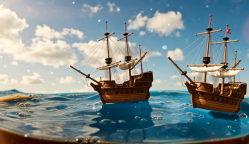} & \includegraphics[width=.2\linewidth]{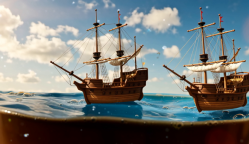} & \includegraphics[width=.2\linewidth]{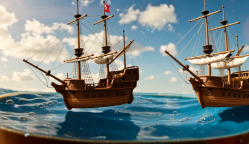} \\
    \end{tabular}
    \caption{Prompt 8: ``\textit{Photorealistic closeup video of two pirate ships battling each other as they sail inside a cup of coffee.}''}
    \label{fig:prompt8}
\end{figure}

\Cref{fig:prompt1}-\ref{fig:prompt8} show the generated videos from different setups. It extracts 4 frames from the generated videos for reference.

%%%%%%%%%%%%%%%%%%%%%%%%%%%%%%%%%%%%%%%%%%%%%%%%%%%%%%%%%%%%%%%%%%%%%%%%%%%%%%%
%%%%%%%%%%%%%%%%%%%%%%%%%%%%%%%%%%%%%%%%%%%%%%%%%%%%%%%%%%%%%%%%%%%%%%%%%%%%%%%

\end{document}